\documentclass[twoside,journal]{IEEEtran}
\usepackage{amsmath,amsfonts}
\usepackage{algorithmic}
\usepackage{algorithm}
\usepackage{array}
\usepackage{textcomp}
\usepackage{stfloats}
\usepackage{url}
\usepackage{verbatim}
\usepackage{graphicx}
\usepackage{cite}
\usepackage{booktabs}
\usepackage{colortbl}
\usepackage{xcolor}
\usepackage{subfigure}
\usepackage{rotating}
\usepackage{multirow}
\usepackage{pifont}
\usepackage[hidelinks]{hyperref}
\begin{document}

\title{3D Segment Anything Model with Visual Mamba for Diagnosing Placenta Accreta Spectrum}
\author{Yuliang Zhang, Fang He, Lulu Peng, Tianyu Yan, Pingping Zhang, Ting Song, Lili Du, Dunjin Chen

\thanks{
Yuliang Zhang, Fang He, Lili Du and Dunjin Chen are with the Department of Obstetrics and Gynecology, The Third Affiliated Hospital, Guangzhou Medical University, Guangzhou 510000, China. (Email: 292260743@qq.com; hefangjnu@126.com; lilidugysy@gzhmu.edu.cn; gzdrchen@gzhmu.edu.cn.)

Fang He is also with the Department of Obstetrics, Guangzhou Women and Children's Medical Center, Guangzhou Medical University, Guangzhou 510180, China.

Lulu Peng and Ting Song are with the Department of Radiology, The Third Affiliated Hospital, Guangzhou Medical University, Guangzhou 510000, China. (Email: 2020683015@gzhmu.edu.cn; flair@gzhmu.edu.cn)

Tianyu Yan and Pingping Zhang are with the School of Future Technology, Dalian University of Technology, Dalian 116024, China. (Email: 2981431354@mail.dlut.edu.cn; zhpp@dlut.edu.cn)
}
}
\markboth{IEEE Transactions on Image Processing}{}
\maketitle
\begin{abstract}
Placenta Accreta Spectrum (PAS) is a rare but highly dangerous obstetric disease.
Early and accurate PAS diagnosis is critical for maternal health.
Traditional PAS diagnosis relies on experienced doctors by analyzing the cesarean history and Magnetic Resonance Imaging (MRI) data.
However, district-level hospitals often lack the expertise and resources for accurate PAS diagnosis.
To address these challenges, we establish the first MRI-based PAS dataset, which includes both fine-grained segmentation and classification annotations.
Meanwhile, diagnosing PAS can be significantly enhanced by segmenting lesion areas from MRI images of the uterus.
To achieve automatic PAS diagnosis, we propose 3DSAMba, a novel feature learning framework for effective lesion segmentation.
More specifically, we first design a 3D Segment Anything Model (SAM) and incorporate medical domain information into the model through an efficient adapter mechanism.
In addition, we introduce a Multi-Level Aggregation Mamba (MLAM) to aggregate feature maps across different levels and a Fusion State Space Model (FSSM) to fuse multi-scale features from both the encoder and decoder.
Finally, we apply segmentation masks to the original MRI images through element-wise multiplication, effectively isolating lesion areas for more accurate PAS diagnosis.
Extensive experiments validate that our framework significantly improves the PAS diagnostic performance.
To facilitate further research in PAS diagnosis, we have released the dataset and source code at https://github.com/Drchip61/PASD.
\end{abstract}

\begin{IEEEkeywords}
Placenta Accreta Spectrum, Segment Anything Model, Vision State Space Model, Multi-Level Feature Fusion.
\end{IEEEkeywords}

\section{Introduction}
\label{sec:introduction}
Placenta Accreta Spectrum (PAS) is a complex obstetric disease that affects the structural interface between the outer placental wall and the inner uterine wall.
It is characterized by the abnormal placental attachment to the uterine wall, which may invade surrounding tissues and organs~\cite{jauniaux2018placenta,poljak2023placenta}.
Depending on the invasion depth of trophoblasts in the uterine myometrium, PAS is classified into three types: (1) placenta accreta, where trophoblasts attach directly to the myometrium without invasion; (2) placenta increta, where trophoblasts invade the myometrium reaching its outer layer; and (3) placenta percreta, where trophoblasts penetrate or pass through the uterine serosa.
The primary concern of diagnosing PAS lies in the difficulty of placenta separation during delivery, which can lead to severe postpartum hemorrhages and even pose a life-threatening risk to the mother.

Currently, diagnosing PAS requires doctors to consider diverse factors, such as abnormal placental signals, blurred boundaries between the placenta and the uterus, heterogeneous enhancement within the placenta, and evidence of placental invasion into the bladder or other surrounding organs.
This requires experienced doctors, making it challenging to achieve effective diagnostics in district-level hospitals.
Therefore, automatic PAS diagnosis is highly necessary.
However, there is no large-scale PAS dataset currently.
Meanwhile, Magnetic Resonance Imaging (MRI) is the most suitable modality for PAS diagnosis~\cite{romeo2019machine} due to its superior soft tissue contrast and its ability to clearly visualize the anatomical relationships between the placenta, myometrium, and surrounding organs.
To this end, we present the first MRI-based PAS dataset, annotated by diverse doctors with different clinical experiences.

Moreover, PAS is primarily characterized by abnormal placental adhesion or invasion into the uterine myometrium.
However, the abundant content in MRI makes accurate diagnosis challenging.
Consequently, precise lesion segmentation can enhance the diagnostic capability.
As for lesion segmentation, many methods based on Convolutional Neural Networks (CNNs) have achieved significant success~\cite{ronneberger2015u,dou20163d,gibson2018automatic,li2018h,yu2017volumetric}.
However, the main limitation of CNNs lies in their local receptive field, which restricts the ability to capture long-range dependencies and global contexts.
To address aforementioned limitations, Transformer-based methods~\cite{vaswani2017attention,khan2022transformers} have been introduced, which capture global contexts.
However, Transformer-based methods require large-scale annotated datasets to fully deliver their potential.
This data scarcity issue makes it difficult for Transformer-based methods to generalize effectively.
Recently, Segment Anything Model (SAM)~\cite{kirillov2023segment} is proposed and has a strong ability of universal image understanding.
However, SAM is primarily trained on natural images, thus it has limited prior knowledge of the medical domain.
Additionally, SAM is originally designed for processing 2D images, and it requires an adaptation for lesion segmentation in 3D MRI data.

Recently, State Space Models (SSMs)~\cite{guefficiently}, particularly Mamba~\cite{gu2023mamba}, have garnered significant attention for their ability to model long-range dependencies while maintaining linear computational complexity.
For example, VMamba~\cite{liu2024vmamba} applies the principles of SSMs to computer vision.
By scanning input data in four directions, VMamba has demonstrated outstanding performance in various perception tasks, including medical image segmentation.
Despite the impressive segmentation results, Mamba lacks the inherent ability to adequately extract multi-level features.
This limitation becomes significant in complex PAS diagnosis, which requires both fine-grained details and long-range contextual information.
To address this issue, we propose to enhance Mamba's ability by incorporating multi-level feature aggregation.
It can better capture both local and global contextual information for PAS lesion segmentation.

To address the above issues, we propose a novel feature learning framework called 3DSAMba to effectively segment the lesion areas for PAS diagnosis.
More specifically, we first design a 3D SAM and integrate medical domain information into the model through an efficient adapter mechanism.
Subsequently, we employ a Multi-Level Aggregation Mamba (MLAM) to aggregate feature maps from different levels.
Then, we utilize a Fusion State Space Model (FSSM) to fully fuse the multi-scale feature maps obtained from both the encoder and decoder.
Through the above modules, we obtain the segmentation masks of PAS lesion areas.
Finally, we use a simple classification network to analyze the lesion areas and achieve an accurate PAS diagnosis.

Our contributions are summarized as follows:
\begin{itemize}
\item
We present the first large-scale PAS diagnosis dataset with both fine-grained segmentation and classification annotations, advancing the obstetric disease research.
\item
We propose a novel feature learning framework called 3DSAMba for effective lesion segmentation to improve PAS diagnosis.
Our framework fully combines the strengths of SAM and Mamba, achieving outstanding results in PAS lesion segmentation.
\item
We propose the Multi-Level Aggregation Mamba (MLAM) and Fusion State Space Model (FSSM), which leverage multi-level and multi-scale features to improve the accuracy of PAS diagnosis.
\item
We perform extensive experiments to verify the effectiveness of the proposed methods.
Our framework can accurately segment PAS lesion areas and obtain better PAS diagnosis results.
\end{itemize}
%-------------------------------
\section{Related Work}
\subsection{Placenta Accreta Spectrum Diagnosis}
Technically, PAS diagnosis is primarily based on three kinds of approaches.
The first kind of approaches are based on the history of cesarean sections~\cite{american2018placenta,silver2018placenta}.
Statistical analysis reveals a clear correlation between the number of cesarean deliveries and the incidence of PAS.
Specifically, women who have undergone more than three cesarean sections are at a significantly higher risk of developing PAS.
Although this kind of approaches are simple and direct, they have the lowest reliability~\cite{arakaza2023placenta}.
The second kind of approaches involve ultrasound examination~\cite{miller2021placenta,baughman2008placenta,happe2021predicting}.
In ultrasound imaging, certain imaging features are considered for the indicative PAS risk.
These features include an unclear boundary between the placenta and the uterine myometrium, thinning or interruption of the myometrium, and increased blood flow within the placenta.
However, the ultrasound examination mainly relies on echo patterns.
It can be affected when the placenta is located on the posterior uterine wall.
The third kind of approaches are based on MRI analysis~\cite{srisajjakul2020magnetic,ishibashi2020use,kapoor2021review}.
MRI provides superior soft tissue contrast and allows for a more precise assessment of placental invasion depth.
T2-weighted Imaging (T2WI) is particularly effective in visualizing anatomical relationships among the placenta, uterine myometrium, bladder, and adjacent organs.
This makes MRI especially valuable for identifying posterior placental implantation and other complex PAS cases.
Previous studies~\cite{de2022diagnosis,do2020mri} have demonstrated that MRI-based PAS diagnosis depends on several key imaging features.
However, identifying these MRI features is very challenging for doctors in district-level hospitals.
\subsection{CNN-based Medical Image Segmentation}
In recent years, CNN-based methods have achieved great advancement in medical image segmentation~\cite{ronneberger2015u,dou20163d,gibson2018automatic,li2018h,yu2017volumetric}.
For example, Ronneberger \textit{et al}.~\cite{ronneberger2015u} propose the first CNN-based encoder-decoder structure named U-Net for medical image segmentation.
Zhou \textit{et al}.~\cite{zhou2018unet++} further introduce dense skip pathways into the encoder-decoder structure for complex medical image segmentation.
Mortazi \textit{et al}.~\cite{mortazi2018automatically} propose an automatically designing CNN method for medical image segmentation.
Chen \textit{et al}.~\cite{chen2018drinet} integrate U-Net with Atrous Spatial Pyramid Pooling (ASPP) to enhance semantic and contextual feature extraction.
Chen \textit{et al}.~\cite{chen2018encoder} propose an effective CNN decoder to refine the segmentation results along lesion boundaries.
Jha \textit{et al}.~\cite{jha2019resunet++} propose a residual CNN for colonoscopic image segmentation.
Gu \textit{et al}.~\cite{gu2019net} propose a context-aware CNN to capture more high-level information and preserve spatial information.
Jha \textit{et al}.~\cite{jha2020doubleu} introduce densely connected CNN for medical image segmentation.
Baldeon \textit{et al}.~\cite{baldeon2020adaresu} propose a multi-objective adaptive CNN for medical image segmentation.
Adegun \textit{et al}.~\cite{adegun2021deep} propose an enhanced U-Net for medical image analysis, capable of segmenting and identifying lesion areas.
Valanarasu \textit{et al}.~\cite{valanarasu2022unext} propose a light-weight CNN for rapid medical image segmentation.
However, due to the limited receptive fields, CNN-based methods lack the ability to model long-range dependencies.
%-------------------------------
\begin{figure*}[!t]
\centerline{\resizebox{0.84\textwidth}{!}{
  \includegraphics{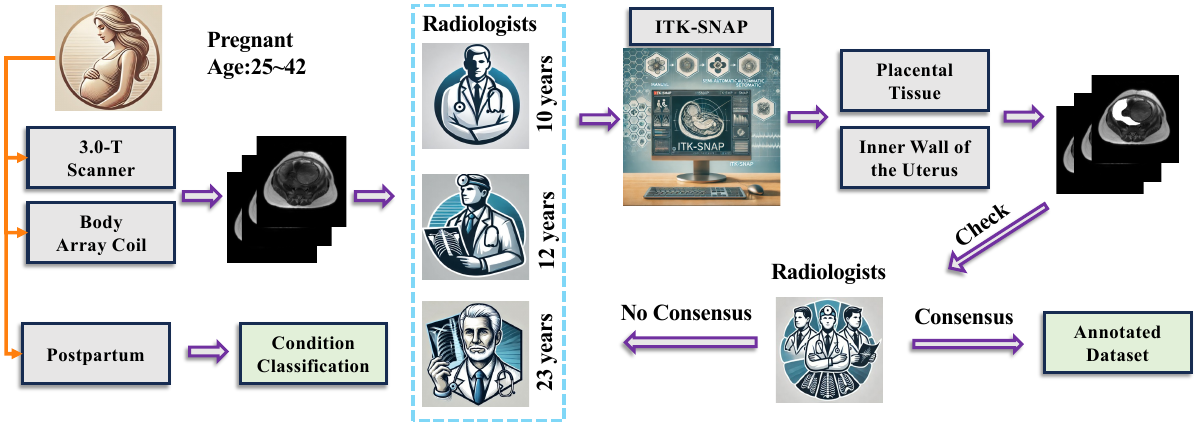}}}
\caption{The pipeline of our data acquisition and annotation.}
\vspace{-4mm}
\label{dataset}
\end{figure*}
%---------------------------------------------
\begin{table*}[h!]
\centering
\caption{Demographic characteristics and clinical covariates of patients in the proposed PAS dataset.}
\small
\resizebox{0.84\textwidth}{!}{
\begin{tabular}{l|c|c|c|l}
\hline
 Demographic Characteristics and Clinical Covariates &Level	&Non-PAS	&PAS&	p-value\\
 \hline
Maternal age [years, mean ± SD]	&&	33.60 (5.46)&	33.57 (4.12)&	0.962\\
Gestational age at MRI scan [weeks, mean ±  SD]&	&	32.27 (5.49)&	32.52 (5.42)	&0.735\\
\hline
Pregnancy history&	Negative [number (percentage)]	&7 (8.5)&	1 (0.6)&	0.004\\
	&Positive [number (percentage)]	&	75 (91.5)	&161 (99.4)	&\\
    \hline
Abortion history	&Negative [number (percentage)]		&28 (34.1)	&37 (22.8)&	0.083\\
	&Positive [number (percentage)]	&	54 (65.9)	&	125 (77.2)	&\\
    \hline
Previous cesarean delivery and/or curettage	&Negative [number (percentage)]		&27 (32.9)&	8 (4.9)	&$<$0.001\\
	&Positive [number (percentage)]	&	55 (67.1)	&154 (95.1)	&\\
    \hline
Placenta previa&	Negative [number (percentage)]		&28 (34.1)	&16 (9.9)	&$<$0.001\\
	&Positive [number (percentage)]	&	54 (65.9)	&146 (90.1)	&\\

\hline
\end{tabular}
}
\label{table:data_distribution}
\end{table*}
\subsection{Transformer-based Medical Image Segmentation}
Inspired by the success of Vision Transformers (ViT)~\cite{alexey2020image,liu2021swin}, Xie \textit{et al}.~\cite{xie2021cotr} combine CNN with Transformer to effectively capture long-range dependencies in medical image segmentation.
Chen \textit{et al}.~\cite{chen2021transunet} further combine global context extraction through Transformers with precise localization via CNNs.
Hatamizadeh \textit{et al}.~\cite{hatamizadeh2022unetr} combine a Transformer encoder and a CNN decoder with a U-shaped network design, and use skip connections to link the encoder and decoder for precise 3D medical image segmentation.
Furthermore, Hatamizadeh \textit{et al}.~\cite{hatamizadeh2021swin} leverage a hierarchical Swin Transformer~\cite{liu2021swin} to extract multi-resolution features for more precise segmentation.
Recently, SAM~\cite{kirillov2023segment} is proposed to achieve an exceptional image understanding capability.
It also performs well in medical domains, particularly in data-scarce medical image segmentation tasks~\cite{chen2024ma,gong20243dsam,zhang2023input}.
However, SAM is originally designed to process 2D images.
The inherent challenges of 3D medical images make it difficult for SAM to fully deliver its capabilities.
Some work has made efforts to incorporate SAM into 3D medical image segmentation.
For example, Bui \textit{et al}.~\cite{bui2024sam3d} directly process 3D medical images by treating them as individual 2D slices.
Gong \textit{et al}.~\cite{gong20243dsam} introduce spatial adapters for fine-tuning, allowing SAM to effectively capture spatial patterns in medical images with minimal adjustments.
Li \textit{et al}.~\cite{li2023auto} develop a prompt learning method to adapt SAM for 3D medical image segmentation.
However, since SAM is originally designed for 2D data, previous methods may result in suboptimal performance.
\subsection{Mamba-based Medical Image Segmentation}
State Space Models (SSMs)~\cite{gu2023mamba,wang2023selective} offer a promising approach for efficiently processing long sequences.
Recently, based on SSMs, Mamba has a strong potential in improving the accuracy of segmenting medical images.
For example, Ma \textit{et al}.~\cite{ma2024u} propose a universal medical image segmentation architecture called U-Mamba.
Wang \textit{et al}.~\cite{wang2024mamba} combine Mamba with skip connections to preserve spatial information across different feature scales.
Wang \textit{et al}.~\cite{wang2024lkm} design a large kernel-based vision Mamba for medical image segmentation.
Zhang \textit{et al}.~\cite{zhang2024vm} propose a semantic and detail infusion module with Mamba to enhance the low-level and high-level features.
Xing \textit{et al}.~\cite{xing2024segmamba} propose a long-range sequential modeling Mamba for 3D medical image segmentation.
Liu \textit{et al}.~\cite{liu2024swin} couple Mamba and Swin Transformer for medical image segmentation.
Wu \textit{et al}.~\cite{wu2025h} propose a high-order vision Mamba U-Net for medical image segmentation.
Shi \textit{et al}.~\cite{shi2025frequency} propose a frequency-enhanced Mamba network for efficient vertebrae segmentation.
Wang \textit{et al}.~\cite{wang2026comprehensive} provide a comprehensive analysis of Mamba for 3D volumetric medical image segmentation, highlighting the potential and challenges of Mamba-based architectures in this domain.
Although achieving outstanding segmentation results, these Mamba-based methods fail to effectively capture the spatial correlations of multi-level features.
\section{The Placenta Accreta Spectrum Dataset}
Early and accurate PAS diagnosis is critical for ensuring maternal safety.
However, there is still a lack of systematic research on PAS diagnosis.
An important reason is that there is no publicly available PAS dataset.
To this end, we first build a high-quality MRI-based PAS dataset, which can provide a solid foundation for PAS diagnosis.
Fig.~\ref{dataset} illustrates the pipeline of our data collection and annotation.
\subsection{MRI Data Acquisition}
The pelvic MRI data is acquired by using a 3.0-T scanner equipped with a 32-channel body array coil.
The imaging protocol focuses on T2-weighted Imaging (T2WI) to ensure an optimal visualization of the placenta and uterine structures.
The patients are scanned in the supine position, and all images are classified according to the clinical diagnosis after delivery.

The dataset comprises MRI scans of 244 patients, including 162 positive PAS cases and 82 negative PAS cases.
The maternal age ranges from 25 to 42 years.
The gestational age at the time of MRI scan is consistent across the groups, averaging 33 weeks.
As shown in Tab.~\ref{table:data_distribution}, continuous variables such as maternal age and gestational age at MRI are compared using independent t-tests, while discrete variables such as pregnancy history, cesarean section history, and placenta previa status are evaluated using Chi-square tests.
No significant differences are observed in maternal age, gestational age at MRI, or abortion history (p $>$ 0.05).
However, significant group differences (p $<$ 0.05) are found in pregnancy history, cesarean section history, and placenta previa status.
These variables are potential confounders that may influence model predictions and have been considered when interpreting the classification results.
This PAS dataset ensures comprehensive representations of PAS scenarios, enabling robust model development and evaluation.
%----------------------------------------------------
\begin{figure}[t]
\centerline{\resizebox{0.46\textwidth}{!}{\includegraphics{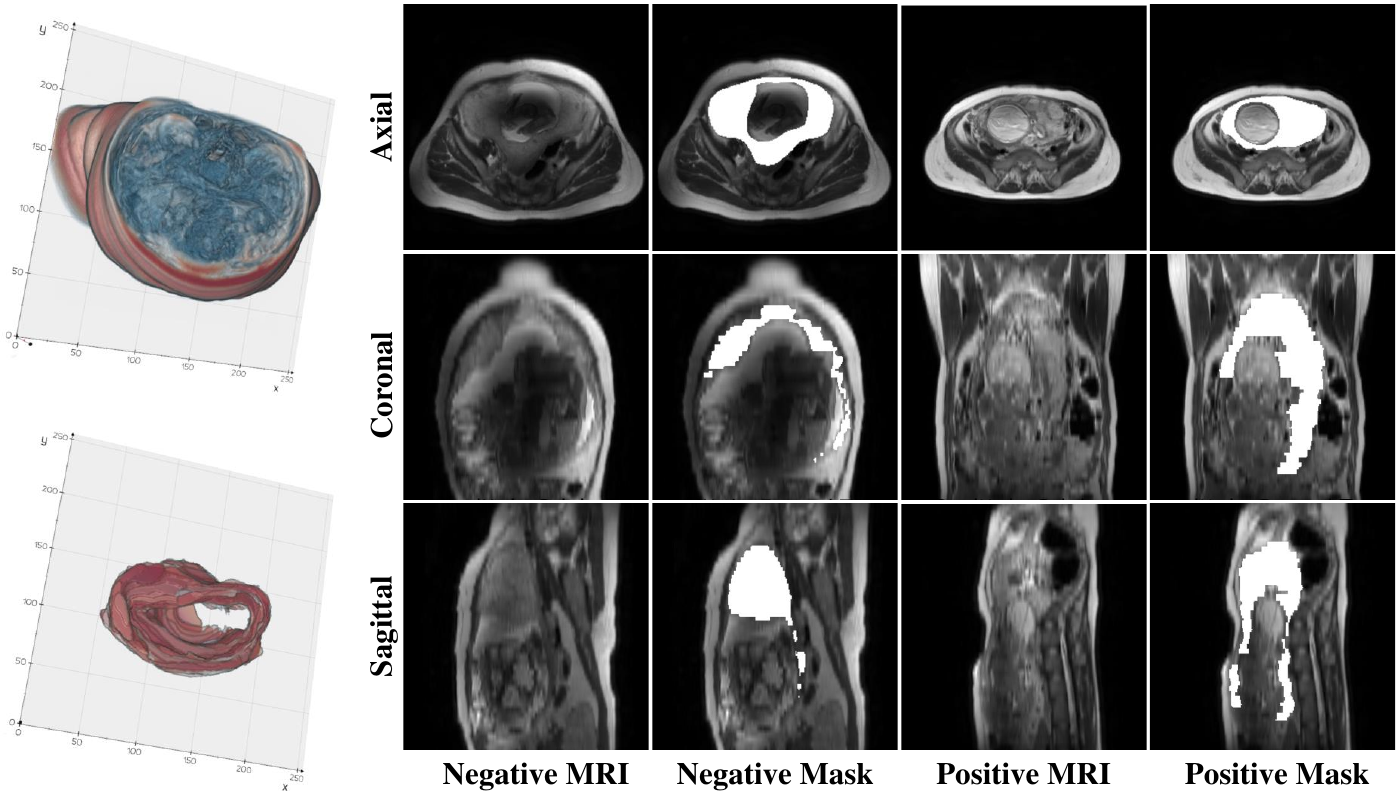}}}
\caption{Visualization of the original MRI data and PAS lesion areas in both 3D volumes and 2D slices.}
\vspace{-2mm}
\label{motivation}
\end{figure}
\subsection{Data Annotation}
The annotations are collaboratively obtained by three radiologists with 10, 12, and 23 years of experiences, respectively.
Based on T2WI images, the radiologists manually segment the outer placental wall and the inner uterine wall with the aid of ITK-SNAP (version 3.6.0; www.itksnap.org), a widely used tool for medical image segmentation.
To ensure the quality, each radiologist independently performs segmentation on the same cases.
The segmentation results are then cross-reviewed, and any discrepancies are resolved through joint discussions to reach a consensus.
This standardized multi-review procedure ensures high-quality and consistent annotations, providing a reliable foundation for PAS analysis.

\begin{figure*}[t]
\centering
\includegraphics[width=0.95\textwidth]{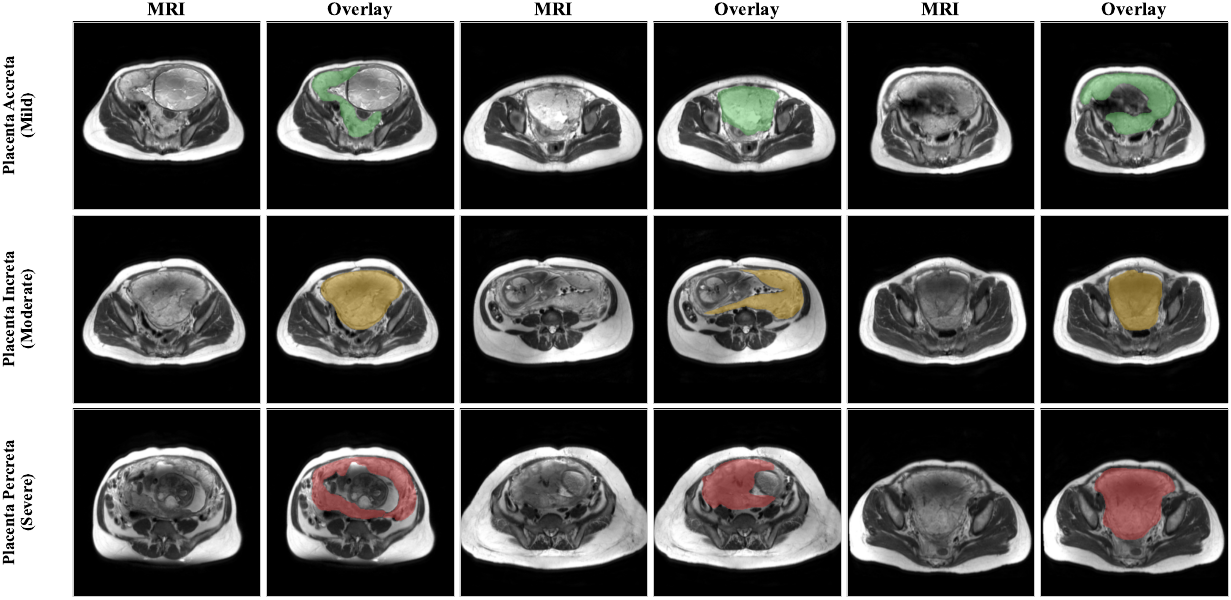}
\caption{Representative MRI slices and lesion overlays for the three PAS subtypes stratified by quantitative severity. Green, orange, and red overlays denote accreta (mild), increta (moderate), and percreta (severe), respectively. Each row shows three representative cases with increasing invasion severity.}
\label{subtypes}
\end{figure*}

\begin{figure}[t]
\centering
\includegraphics[width=0.46\textwidth]{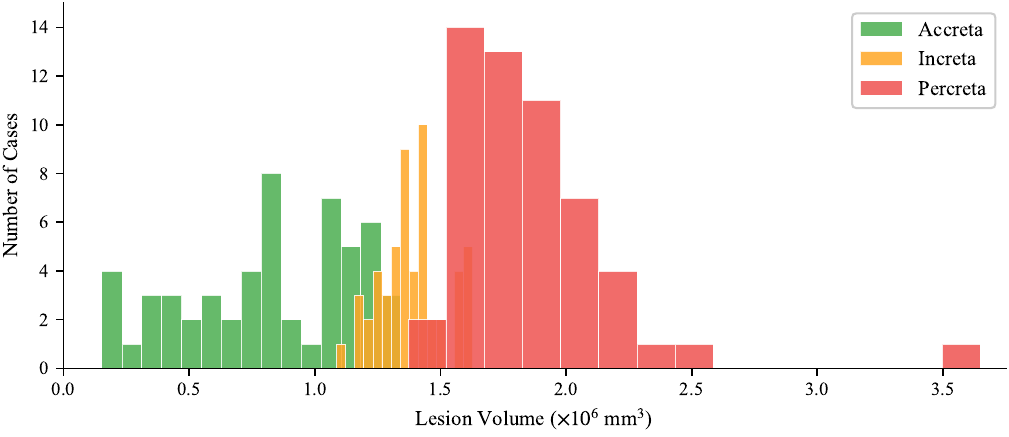}
\caption{Distribution of lesion volumes across three PAS severity groups.}
\label{subtypes_dist}
\end{figure}

In Fig.~\ref{motivation}, the left column shows 3D visualizations of the original MRI data and the lesion areas.
The right panel displays representative 2D slices from both negative and positive cases in axial, coronal, and sagittal planes.
The first and third columns present the original MRI images, while the second and fourth columns show the corresponding lesion masks overlaid in white.
In negative cases, the interface between the outer wall of the placenta and the inner wall of the uterus is clear, and no abnormal invasions are found.
In contrast, positive cases exhibit characteristic features of PAS subtypes.
The placenta accreta shows trophoblasts attaching to the myometrium without invasion, the placenta increta shows trophoblasts invading the myometrium with partial disruptions of the uterine contour, and the placenta percreta shows trophoblasts penetrating the uterine serosa and extending into adjacent tissues.
These representative examples highlight the diagnostic features captured in the dataset and provide a visual reference for understanding PAS presentations.
Our dataset provides binary-level annotations (i.e., PAS-positive vs. PAS-negative) rather than explicit subtype labels (accreta, increta, percreta).
This is because fine-grained subtype differentiation on MRI alone remains clinically challenging, as the definitive subtype diagnosis often relies on intraoperative findings and postpartum histopathology.
To further characterize the severity distribution within the positive cases, we compute a composite severity score for each case based on the lesion volume, the proportion of affected slices, and the maximum cross-sectional lesion area.
Using this score, we divide the 162 positive cases into three groups corresponding to the PAS spectrum: accreta, increta, and percreta.
As shown in Fig.~\ref{subtypes}, representative examples from each group clearly demonstrate increasing invasion extent: accreta cases exhibit relatively small and focal lesions, increta cases show moderate invasion with larger lesion volumes, and percreta cases display extensive invasion with the largest lesion coverage.
Fig.~\ref{subtypes_dist} further presents the lesion volume distribution across the three groups, confirming that our dataset spans a wide range of invasion severities and naturally encompasses the full PAS spectrum.
In the future, incorporating pathology-confirmed subtype annotations will enable more fine-grained subtype-level analysis.
\section{Proposed Method}
\begin{figure*}[h]
\centerline{\resizebox{0.9\textwidth}{!}{\includegraphics{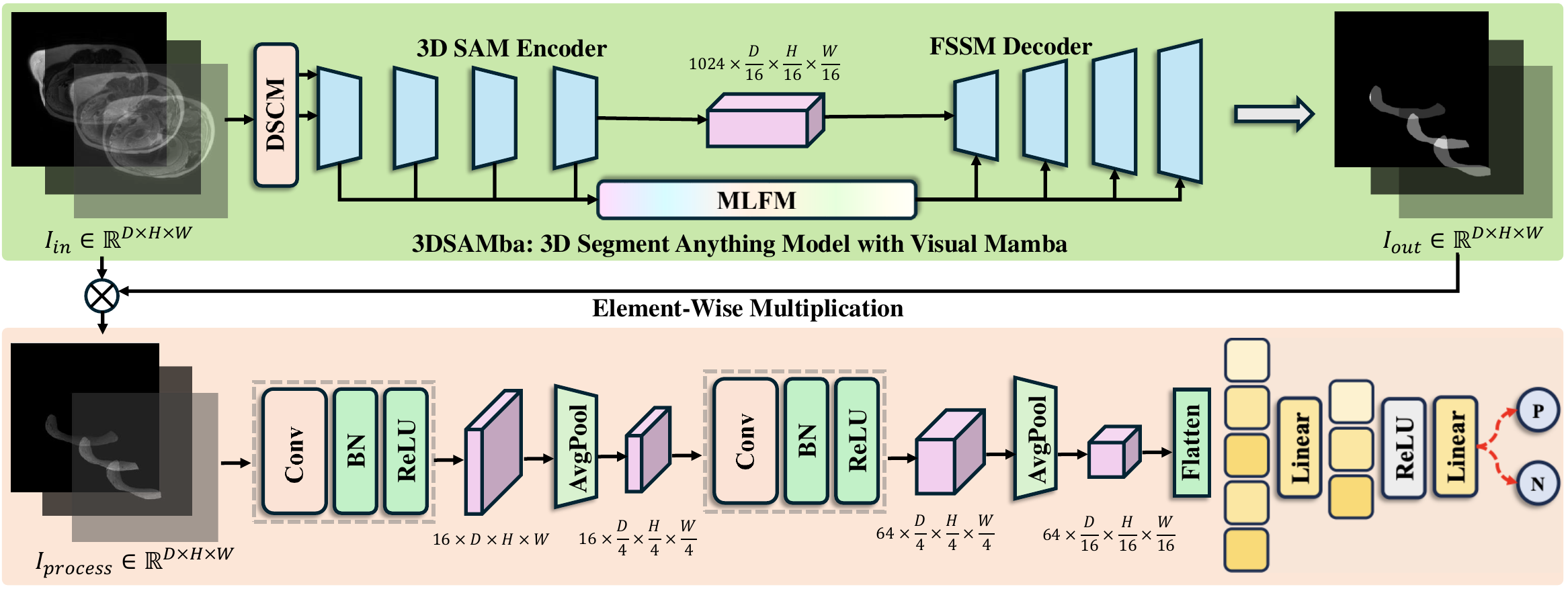}}}
\caption{Illustration of our proposed framework for PAS diagnosis. It includes a lesion segmentation model and a simple classification network to localize the lesion areas and achieve PAS diagnosis.}
\label{framework}
\end{figure*}

\begin{figure*}[!t]
\centerline{\resizebox{0.9\textwidth}{!}{\includegraphics{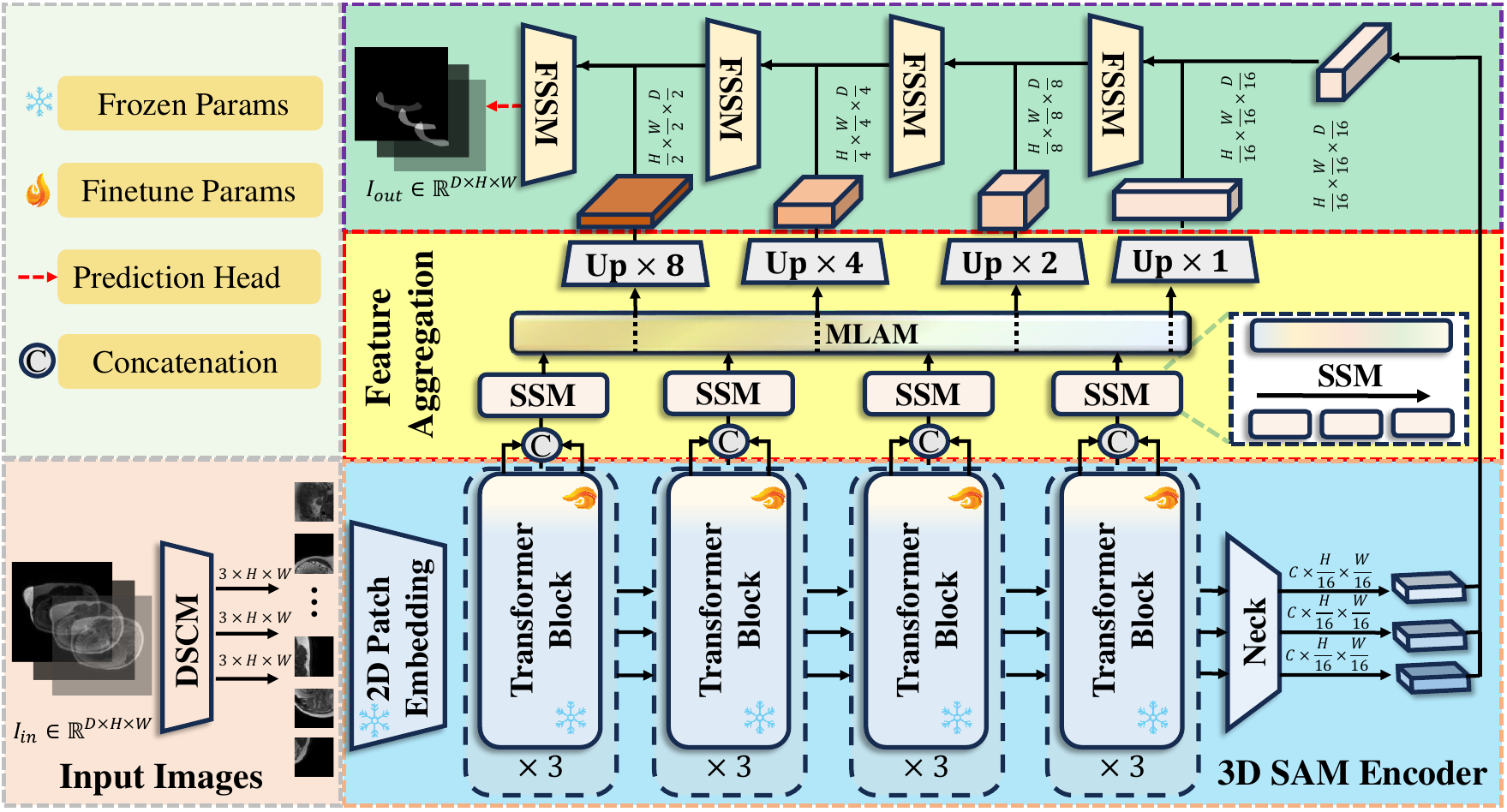}}}
\caption{Illustration of our proposed 3DSAMba. It includes three main components: 3D Segment Anything Model (SAM) Encoder, Multi-Level Aggregation Mamba (MLAM) and Fusion State Space Model (FSSM).}
\label{3DSAM}
\end{figure*}

As shown in Fig.~\ref{framework}, our method firstly segments lesion areas by 3DSAMba.
Then, we perform an element-wise multiplication with the segmentation masks and input MRI images.
Finally, we use a simple classification network to achieve PAS diagnosis.
As shown in Fig.~\ref{3DSAM}, our 3DSAMba includes three main components: 3D Segment Anything Model (SAM) Encoder, Multi-Level Aggregation Mamba (MLAM) and Fusion State Space Model (FSSM).
These key components are elaborated in the following sections.
%---------------------------------------
%--------------------------------------------------------------------------------------
\subsection{3D SAM Encoder}
SAM is originally designed for 2D image segmentation and lacks the inherent ability to directly handle the spatial complexities of 3D volumetric data.
Previous attempts to apply SAM to 3D data struggle to preserve essential spatial information in volumetric inputs.
A straightforward alternative is to modify SAM into a native 3D model by replacing all 2D operators with their 3D counterparts.
However, such a strategy has three main drawbacks:
(1)~it breaks the compatibility with SAM's large-scale 2D pre-trained weights, forcing a full retraining on the target domain where annotated 3D medical data are scarce;
(2)~3D self-attention incurs a cubic memory and computation growth with respect to the spatial resolution, making it prohibitive for high-resolution MRI volumes;
and (3)~the size of our PAS dataset (244 cases) is insufficient to train a 3D vision foundation model from scratch without severe overfitting.
To retain the powerful pre-trained representations of SAM while enabling 3D volumetric understanding, we instead adopt a \emph{compress-then-process} paradigm: a lightweight learnable module transforms the 3D input into a pseudo-2D representation that the frozen SAM encoder can directly process.
To address these limitations, we propose a novel Deep Sequence Compression Module (DSCM) for 3D data.
It aims to transform 3D data into a format that simulates 2D inputs, making it compatible with the SAM architecture.
This module compresses the volumetric input into a 9-channel representation, effectively reducing its dimension while preserving essential spatial information.
Then, the compressed representation is further divided into three separated 3-channel slices.
In this manner, we can effectively process 3D inputs by the frozen SAM architecture.
This transformation ensures that the abilities of SAM can be fully utilized while maintaining the compatibility with 3D medical data.
Specifically, the DSCM is expressed as follows:
\begin{equation}
I^{1}, I^{2}, I^{3}=\operatorname{Split}\left(\gamma\left(\theta\left(\psi\left(I_{in}\right)\right)\right)\right),
\end{equation}
where $I_{\text {in }} \in \mathbb{R}^{D \times H \times W}$ is the 3D input data.
$D$, $H$ and $W$ denote the depth, height and width dimensions, respectively.
$I^{\text {1,2,3 }} \in \mathbb{R}^{3 \times H \times W}$ is the actual input of SAM's backbone.
$\psi$ represents a 3D convolution, which compresses the initial MRI input into 9 channels.
$\theta$ denotes the Batch Normalization (BN)~\cite{ioffe2015batch}, and
$\gamma$ represents the Rectified Linear Unit (ReLU)~\cite{li2017convergence}.
Through DSCM, we obtain 2D data that can be manageable by SAM.

\begin{figure}[h]
\centerline{\includegraphics[width=0.4\textwidth]{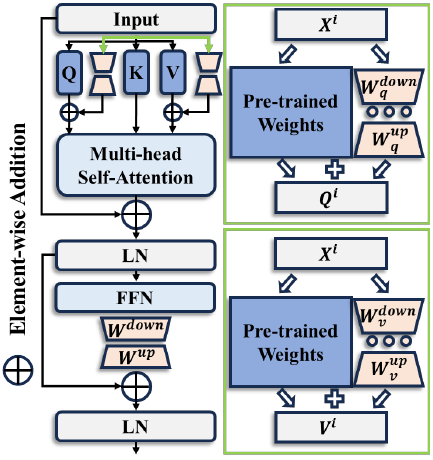}}
\caption{Illustration of our proposed adapters in each Transformer block.}
\label{adapter}
\end{figure}

While SAM demonstrates exceptional segmentation abilities, its training data is mainly from natural images.
Consequently, SAM lacks the domain-specific knowledge to fully understand the complex structural patterns and nuances present in medical images.
This discrepancy poses a significant limitation when applying SAM to medical scenarios.
To bridge this gap, we augment SAM's frozen backbone with a low-rank adapter, as shown in Fig.~\ref{adapter}.
This design allows SAM to adapt its pre-trained features to medical contexts and maintain the general visual understanding~\cite{hu2022lora,zhang2024fantastic}.
By selectively introducing medical domain information, the adapter effectively enhances SAM's ability to extract meaningful features from medical images.
This approach preserves the advantages of SAM while tailoring it to the unique characteristic of medical images.
More specifically, the detailed formulations are expressed as follows:
\begin{equation}
Q_{i}=\phi\left(X_{i}\right) W_{q}+\left(\phi\left(X_{i}\right) W_{q}^{\text{down}}\right) W_{q}^{\text{up}},
\end{equation}
\vspace{-2mm}
\begin{equation}
V_{i}=\phi\left(X_{i}\right) W_{v}+\left(\phi\left(X_{i}\right) W_{v}^{\text{down}}\right) W_{v}^{\text{up}},
\end{equation}
\vspace{-2mm}
\begin{equation}
S_{i}=\operatorname{Softmax}\left(\frac{Q_{i} K_{i}^{T}}{\sqrt{d}}\right) V_{i},
\end{equation}
\vspace{-2mm}
\begin{equation}
X_{i+1}=\gamma\left(\eta\left(\phi\left(S_{i}\right)\right) W^{\text{down}}\right) W^{\text{up}}+S_{i},
\end{equation}
where $\phi$ is the Layer Normalization (LN)~\cite{ba2016layer} and $\eta$ is the Multilayer Perceptron (MLP)~\cite{almeida2020multilayer}.
$X_{i}\in \mathbb{R}^{T\times D}$ is the input of the $i$-th Transformer block.
$T$ is the number of tokens and $D$ is the embedding dimension.
Moreover, $Q_{i}, K_{i}, V_{i}\in \mathbb{R}^{T\times D}$ are the Query, Key and Value input of self-attention layers, respectively.
$S_{i}\in \mathbb{R}^{T\times D}$ is the hidden state in Transformer blocks.
$W_{q,v}^{\text {down}}\in \mathbb{R}^{D\times f}$ compresses the parameters to a lower dimension, facilitating more efficient learning of medical-specific features, while $W_{q,v}^{\text {up}}\in \mathbb{R}^{f\times D}$ expands back to the original dimension.
Here, $f$ stands for the compressed dimension.
Similarly, $W^{\text{down}}$ and $W^{\text{up}}$ have the same operation with $W_{q}^{\text{down}}$, $W_{v}^{\text{down}}$ and $W_{q}^{\text{up}}$, $W_{v}^{\text{up}}$.
After every three Transformer layers, the outputs of the three branches are concatenated and fed into MLAM.
This low-rank adapter achieves a balance between preserving the features pre-trained in SAM and enabling the use of medical information, ensuring that the model remains robust when adapting to the PAS diagnosis.
The features obtained by the final Transformer layer are passed through SAM's original neck structure.
It compresses the channel dimension to C.
Then, they are concatenated to obtain an input feature of the decoder with a channel dimension $\frac{D}{16}$.
%------------------------------------------
%--------------------------------------------
\begin{figure}[h]
\centerline{\resizebox{0.46\textwidth}{!}{\includegraphics{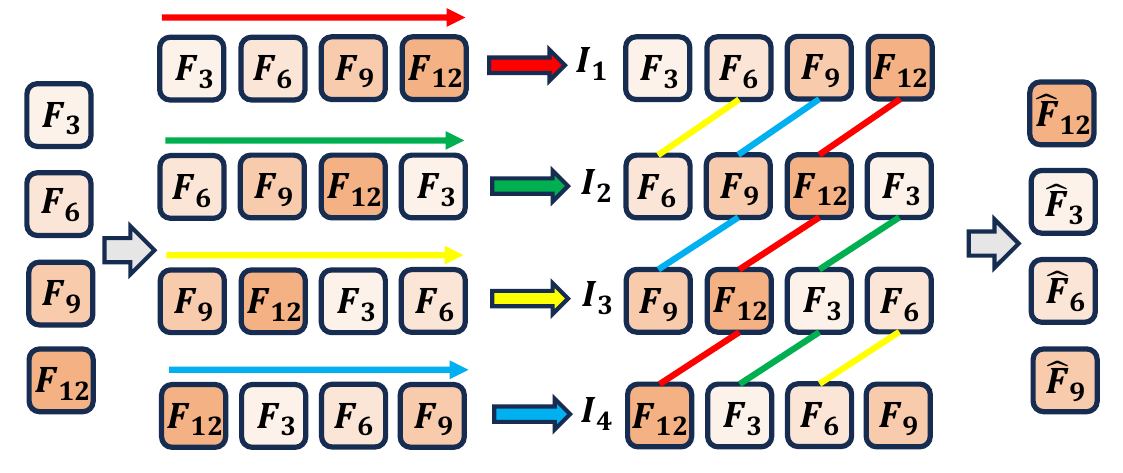}}}
\caption{Illustration of our proposed MLAM.}
\label{mambafussion}
\end{figure}
%--------------------------------------------------
\subsection{Multi-Level Aggregation Mamba}
In SAM, feature maps at different layers capture various levels of semantic and detail information.
Lower layers typically focus on fine-grained details, while higher layers emphasize more semantic contexts.
To better leverage the full range of information, it is essential to aggregate features from different levels.
To this end, we propose the Multi-Level Aggregation Mamba (MLAM), as shown in Fig.~\ref{mambafussion}.
Specifically, we first extract feature maps from the 3-th, 6-th, 9-th, and 12-th Transformer layers of SAM.
Then, we concatenate and reduce the channel dimension to $T$.
Afterwards, we scan these feature maps in different directions, ensuring that the information from each layer is fully interacted with the others.
The procedure is formed as follows:
\begin{equation}
F_{k}=\operatorname{SSM}([X^{1}_{k},X^{2}_{k},X^{3}_{k}]),
\end{equation}
\vspace{-6mm}
\begin{equation}
\begin{array}{l}
I_{1}=\operatorname{SSM}\left([F_{3}, F_{6}, F_{9}, F_{12}]\right), \\
I_{2}=\operatorname{SSM}\left([F_{6}, F_{9}, F_{12}, F_{3}]\right), \\
I_{3}=\operatorname{SSM}\left([F_{9}, F_{12}, F_{3}, F_{6}]\right), \\
I_{4}=\operatorname{SSM}\left([F_{12}, F_{3}, F_{6}, F_{9}]\right),
\end{array}
\end{equation}
where [,] is the concatenation operation.
$SSM$ is the state space model~\cite{gu2023mamba,wang2023selective}
$F_{k}\in \mathbb{R}^{T\times D}$ is the output of the $k$-th Transformer layer and $k\in{\{3,6,9,12\}}$.
$I_{j}\in \mathbb{R}^{4T\times D}$ represents the scanned results from different orders and $j\in{\{1,2,3,4\}}$.
This process ensures that each feature map interacts with others from different layers and positions.
Subsequently, the fused features are aggregated back into their respective positions to maintain the spatial coherence.
The formulation can be represented as:
\begin{equation}
\begin{split}
\hat{F}_{3}=F_{3}+I_{1}[0:T]+I_{2}[3T:4T]\\+I_{3}[2T:3T]+I_{4}[T:2T]\text{,}
\end{split}
\end{equation}
where $\hat{F}_{3}$ is the processed feature map.
The remaining feature maps $\hat{F}_{6}$, $\hat{F}_{9}$, and $\hat{F}_{12}$ are obtained in a similar way.
Through this fusion, we enable information aggregation in multi-level feature maps and make the information at each level more comprehensive.
The whole module enhances the model’s ability to process complex medical images by aggregating low-level details with high-level semantics.
%-----------------------------------------
\begin{figure}[h]
\centerline{\resizebox{0.4\textwidth}{!}{\includegraphics{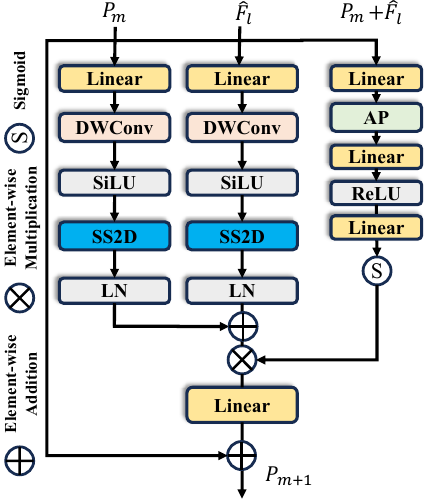}}}
\caption{Illustration of our proposed FSSM.}
\label{fssm}
\end{figure}
%-------------------------------
\subsection{Fusion State Space Model}
Multi-scale representations are of great importance for medical image segmentation.
Hence, the multi-scale decoder is required to achieve precise segmentation of lesion areas~\cite{wu2025h,shi2025frequency}.
However, the original SAM decoder is relatively simplistic and can not adequately capture multi-scale information.
To address this issue, we propose a Mamba-based decoder that incorporates multi-scale features to improve the segmentation accuracy.
As shown in Fig.~\ref{3DSAM}, the core is the Fusion State Space Model (FSSM).
It allows a gradual and efficient fusion of multi-scale features.
The structure of FSSM is shown in Fig.~\ref{fssm}.
The detailed formula is as follows:
\begin{equation}
\bar{F}_{l}=\phi(\operatorname{SS2D}(\operatorname{\beta}(\operatorname{DWConv}(\omega(\hat{F}_{l}))))),
\end{equation}
\begin{equation}
\bar{P}_{m}=\phi(\operatorname{SS2D}(\operatorname{\beta}(\operatorname{DWConv}(\omega(P_{m}))))),
\end{equation}
where $SS2D$ is the 2D state space model~\cite{wang2023selective}.
$\beta$ is the Sigmoid Linear Unit (SiLU)~\cite{nwankpa2021activation}, $\omega$ is a linear layer and $\operatorname{DWConv}$ is the depth-wise convolution~\cite{zhang2019depth}.
$\hat{F}_{l}$ ($l=15-k$) is the output of the MLAM.
$P_{m}$ is the $m$-th stage output of the decoder and $m\in{\{1,2,3,4\}}$.

Furthermore, we enhance the feature interaction among channels.
Formally, this process is expressed as follows:
\begin{equation}
C_{m}= W^{\text{up}}_{c}(\gamma(W^{\text{down}}_{c}(\operatorname{AP}(\omega(\hat{F}_{l}+P_{m}))))),
\end{equation}
where $C_{m}$ is the weight along the channel dimension, and $\text{AP}$ is the average pooling operation.

Once the above weights are obtained, we multiply them to select the key features as follows:
\begin{equation}
P_{m+1} = P_{m}+\omega(\sigma(C_{m})*(\bar{P}_{m}+\bar{F}_{l})),
\end{equation}
where $\sigma$ is the Sigmoid activation function~\cite{pratiwi2020sigmoid}.
This process achieves the feature re-calibration, which adjusts the contribution of each channel based on its relevance to the segmentation task.
The re-calibrated features are then passed through the next stages of the decoder.
The whole process ensures that the final output is enriched with multi-scale information.
\subsection{Loss Function}
To optimize our framework, we use three loss functions: softmax cross-entropy loss (SCE loss), binary cross-entropy loss (BCE loss) and Dice loss.
The SCE loss is used to distinguish positive and negative PAS cases.
The BCE loss is used to measure the voxel-wise classification for segmentation.
The Dice loss focuses on optimizing the overlap between the predicted mask and the ground truth.
The final loss is a combination of these three losses:
\begin{equation}
\mathcal{L} = \mathcal{L}_{\text{SCE}}+\mathcal{L}_{\text{BCE}}+\mathcal{L}_{\text{Dice}}.
\end{equation}
\subsection{Placenta Accreta Spectrum Diagnosis}
In this section, we describe the procedure of PAS diagnosis based on the above lesion segmentation mask.
The motivation behind the \emph{segmentation then classification} strategy is that PAS diagnosis fundamentally depends on the pathological changes at the placenta--uterine interface.
Pelvic MRI images contain abundant anatomical structures beyond the region of interest, including the bladder, bowel loops, amniotic fluid, and pelvic soft tissues.
When a direct classification method is applied to the entire MRI volume, the model may learn spurious correlations from these background structures, leading to unreliable predictions.
Moreover, variations in scan coverage, imaging artifacts, signal intensity inhomogeneity, and noise across different MRI acquisitions can introduce domain shifts that further degrade the performance.
By first segmenting the lesion area, i.e., the region encompassing the outer placental wall and the inner uterine wall, we explicitly constrain the classification input to the pathologically relevant region.
This effectively eliminates interference from irrelevant background structures and reduces the feature space that the classifier needs to learn, enabling more accurate and robust PAS diagnosis.

After our 3DSAMba predicts the lesion areas, we overlay it onto the original MRI image.
As shown in Fig.~\ref{framework}, the lesion areas are retained, while the remaining areas are removed.
After obtaining the processed MRI image, we use a simple 3D CNN for PAS diagnosis.
Specifically, we use two convolutional blocks, each consisting of a 3D convolution with a kernel size of $3 \times 3 \times 3$, followed by a BN and a ReLU activation function.
Each block is followed by an average pooling operation with a kernel size and stride of 4.
The resulting features are then flattened into a vector and passed through a linear layer to predict the confidence of positive and negative PAS cases.
\section{Experiments}
\subsection{Datasets}
In this work, we use two distinct datasets (i.e., the PAS dataset and the KiTS19 dataset~\cite{heller2021state}) to evaluate the performance of our proposed method.
The PAS dataset is proposed by us.
We randomly select 184 cases for training and 60 cases for testing.
In addition, we evaluate our method on the KiTS19 dataset, which is a publicly available CT dataset designed for segmenting kidneys and kidney tumors.
It contains 210 CT scans with detailed annotations.
Following previous works~\cite{heller2021state}, we randomly split the dataset into 170 samples for training and 40 samples for testing.
\subsection{Data Preprocess}
For model training, we perform the following preprocessing steps on the images of our PAS dataset.
\textbf{Truncation:} Voxel values below the 2.5th percentile and above the 97.5th percentile are clipped to reduce the impact of outliers.
\textbf{Normalization:} We apply z-score normalization to standardize the voxel values.
\textbf{Cropping:} Each MRI scan is cropped and resized to a consistent dimension of $48\times 256 \times 256$ to maintain uniformity across the PAS dataset.
\subsection{Evaluation Metrics}
In this work, we utilize different evaluation metrics depending on the nature of datasets and tasks.
On the PAS dataset, we compute the Dice Similarity Coefficient (Dice) and Intersection over Union (IoU) to assess the binary segmentation performance.
As for the PAS diagnosis, we adopt the Overall Accuracy (OA) to evaluate the classification performance.
On the KiTS19 dataset, we use Mean Dice Similarity Coefficient (mDice) and Mean Intersection over Union (mIoU).
Additionally, to facilitate comprehensive comparisons with existing methods on the KiTS19 dataset, we also report the Surface Dice Coefficient (SDC).
By adopting these diverse metrics, we ensure a comprehensive assessment of model performance.
For these metrics, please see the details in~\cite{heller2021state,azad2024medical}.
\subsection{Implementation Details}
We use the PyTorch toolbox and conduct experiments on a single RTX 3090 GPU.
As for the SAM backbone, we use pre-trained SAM-B.
For the rest of our framework, we initialize the parameters randomly.
We use the AdamW optimizer~\cite{diederik2014adam} to update the parameters, setting the learning rate to 0.001 and the weight decay to 0.1.
During training, we reduce the learning rate by a factor of 10 every 20 epochs.
We train the model for a total of 50 epochs.
Due to the constraint of GPU memory, the batch size is set to 1 for each training iteration.
\subsection{Comparisons with the State-of-the-arts}
In this section, we compare the performance of our method with other state-of-the-art methods.
More specifically, we compare with CNN-based methods, such as 3D UNet~\cite{cciccek20163d} and nnUNet~\cite{isensee2021nnu}, as well as Transformer-based methods, including TransUNet~\cite{chen2021transunet}, CoTr~\cite{xie2021cotr}, UNETR~\cite{hatamizadeh2022unetr} and SwinUNETR~\cite{hatamizadeh2021swin}.
Besides, we compare with two recent Mamba-based methods, U-Mamba~\cite{ma2024u} and SegMamba~\cite{xing2024segmamba}.
Furthermore, we compare with SAM-based methods, including AP-SAM~\cite{li2023auto} and 3DSAM-Adapter~\cite{gong20243dsam}.
Both quantitative and qualitative results demonstrate the superiority of our method.
%---------------------------------
\begin{figure*}[h]
\centering
\resizebox{0.8\textwidth}{!}{\includegraphics{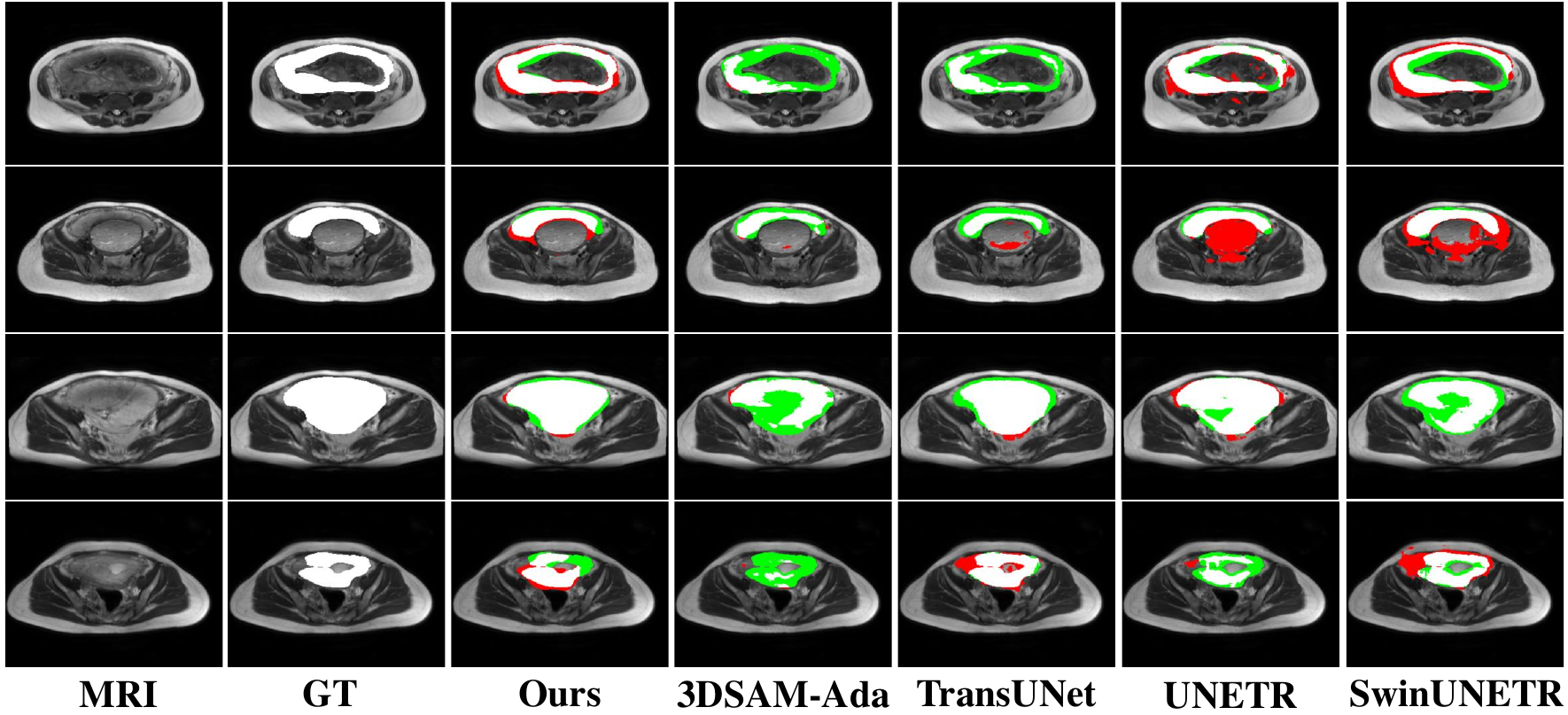}}
\centering
\caption{Visual comparison of predicted masks with different methods in the axial plane. The white areas indicate correctly predicted regions. The red areas represent redundant predictions, while the green areas show missing predictions.}
\label{res_visual1}
\end{figure*}
%-------------------------------------------
\begin{table}[h!]
\centering
\caption{Performance of compared methods on PAS dataset.}
\small
\begin{tabular}{l|c|c|c|c}
\hline
Model &\textbf{Dice} & \textbf{IoU}& \textbf{Params(M)}& \textbf{Time(s)}\\
\hline
3D UNet~\cite{cciccek20163d} & 0.561&0.415&18.16&0.065\\
nnUNet~\cite{isensee2021nnu}&0.570&0.419&19.07&0.066\\
TransUNet~\cite{chen2021transunet}&0.597&0.430&96.07&0.174\\
CoTr~\cite{xie2021cotr}&0.616&0.457&46.51&0.124\\
UNETR~\cite{hatamizadeh2022unetr}& 0.611 & 0.446&92.58&0.078\\
SwinUNETR~\cite{hatamizadeh2021swin} & 0.625&0.478&62.19&0.155\\
U-Mamba~\cite{ma2024u} & 0.495 & 0.334 & 7.50 & 0.130\\
SegMamba~\cite{xing2024segmamba} & 0.647 & 0.518 & 8.44 & 0.070\\
AP-SAM~\cite{li2023auto} & 0.478&0.355&24.14&0.148\\
3DSAM-Ada~\cite{gong20243dsam}&0.525&0.390&13.78&0.130\\
Ours &\textbf{0.728}&\textbf{0.580}&64.68&0.242\\
\hline
\end{tabular}
\label{table:pas_res}
\end{table}
%------------------------------------------------
\begin{table}[h!]
\centering
\caption{Performance of compared methods on KiTS19 dataset.}
\small
\begin{tabular}{l|c|c|c}
\hline
Model &\textbf{mDice} & \textbf{mIoU} & \textbf{SDC}  \\
\hline
TransUNet~\cite{chen2021transunet}& 0.610 & 0.480 & 0.576\\
CoTr~\cite{xie2021cotr}& 0.619&0.495&0.589\\
UNETR~\cite{hatamizadeh2022unetr}  &0.635&0.527&0.607\\
SwinUNETR~\cite{hatamizadeh2021swin}&0.651&0.531&0.617\\
AP-SAM~\cite{li2023auto} & 0.596&0.447&0.531\\
3DSAM-Ada~\cite{gong20243dsam}&0.613&0.487&0.583\\
Ours &\textbf{0.673}&\textbf{0.557}&\textbf{0.634}\\
\hline
\end{tabular}
\label{table:kit_res}
\end{table}
%----------------------------------------------------
\subsubsection{Quantitative Results}
Tab.~\ref{table:pas_res} and Tab.~\ref{table:kit_res} present the experimental results on PAS and KiTS19 datasets, respectively.
As observed, our method outperforms the others across all metrics and demonstrates its superiority.
When compared with CNN-based methods, e.g., 3D UNet~\cite{cciccek20163d} and nnUNet~\cite{isensee2021nnu}, our method outperforms them by approximately 15\% Dice and IoU scores.
This improvement is due to our method’s ability in long-range modeling.
This ability helps our model better capture complex anatomical structures in 3D MRI images and improve the segmentation accuracy.

When compared with Transformer-based methods such as TransUNet~\cite{chen2021transunet}, CoTr~\cite{xie2021cotr}, UNETR~\cite{hatamizadeh2022unetr}, and SwinUNETR~\cite{hatamizadeh2021swin}, our method outperforms them by approximately 10\% IoU and Dice scores on the PAS dataset.
By using the frozen SAM and adapters, our method addresses the data scarcity issue and significantly improves performance.
When compared with Mamba-based methods, i.e., U-Mamba~\cite{ma2024u} and SegMamba~\cite{xing2024segmamba}, our method still achieves better performance on the PAS dataset.
In particular, compared with the strongest SegMamba, our method further improves Dice and IoU by 8.1\% and 6.2\%, respectively.
Finally, we compare our method with SAM-based methods, i.e., AP-SAM~\cite{li2023auto} and 3DSAM-Adapter~\cite{gong20243dsam}.
Our method outperforms them by approximately 20\% in Dice and 19\% in IoU.

To further evaluate the efficiency of our method, we report the number of trainable parameters and the average inference time per case in Tab.~\ref{table:pas_res}.
For SAM-based methods (AP-SAM, 3DSAM-Adapter, and Ours), the reported parameter counts exclude the frozen pre-trained SAM-B backbone (91.10M), since these parameters are shared and not updated during training.
Specifically, CNN-based methods (3D UNet and nnUNet) have relatively small trainable parameter counts (18.16M and 19.07M), while Transformer-based methods range from 46.51M (CoTr) to 96.07M (TransUNet).
SAM-based methods, including AP-SAM (24.14M) and 3DSAM-Adapter (13.78M), have compact trainable modules because the pre-trained SAM backbone is frozen.
Our method has 64.68M trainable parameters, which is still lower than some Transformer-based baselines such as TransUNet (96.07M).
Notably, our method significantly outperforms AP-SAM and 3DSAM-Adapter by approximately 20\% in Dice, indicating that the performance gains are driven by the effective model design rather than merely by the model scale.
While our method introduces a slight increase in computational cost compared to baseline models, it consistently achieves better segmentation performance.
This trade-off is acceptable given the substantial improvements in accuracy, boundary adherence, and lesion completeness.
\begin{figure*}[h]
\centering
\resizebox{0.8\textwidth}{!}{\includegraphics{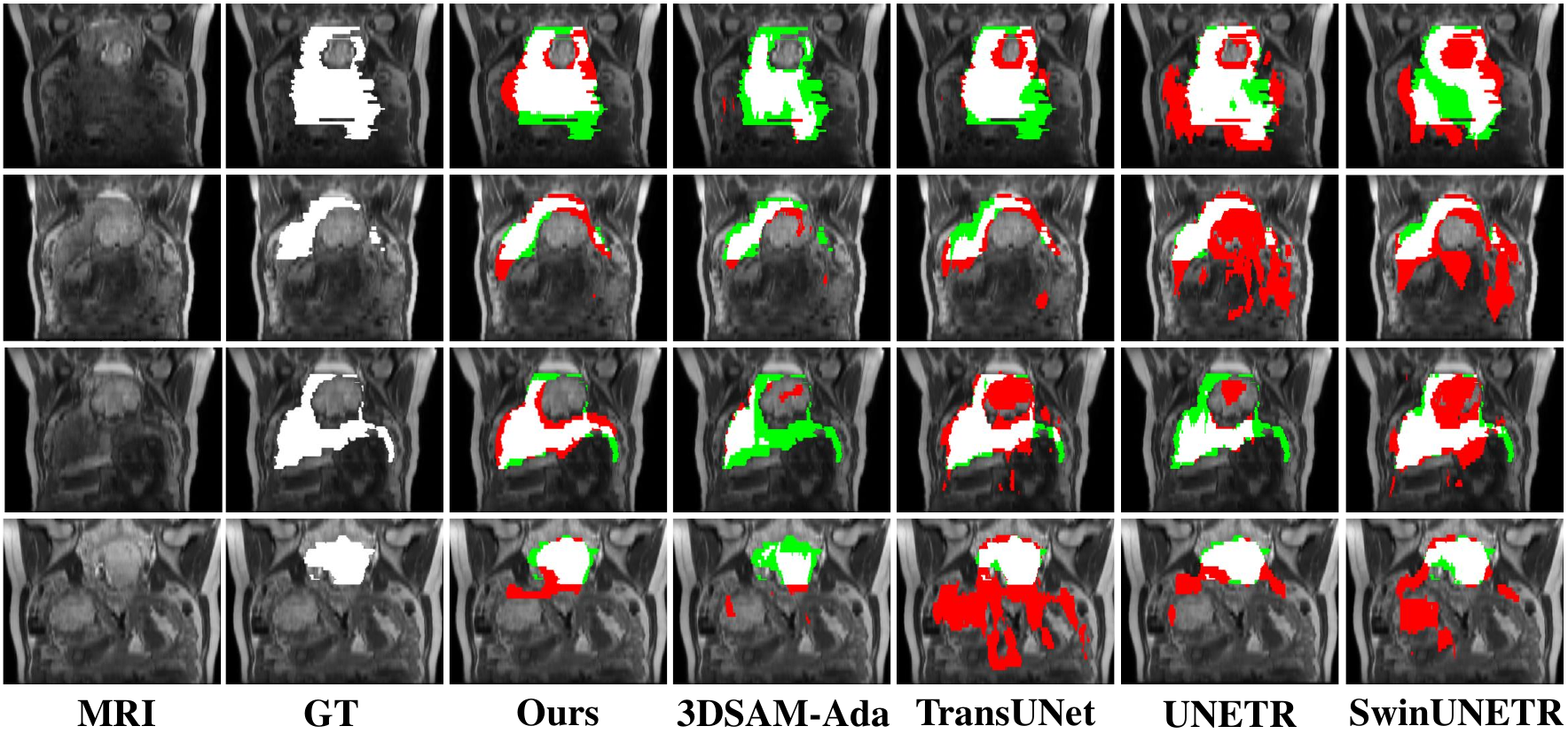}}
\centering
\caption{Visual comparison of predicted masks with different methods in the coronal plane. The white areas indicate correctly predicted regions. The red areas represent redundant predictions, while the green areas show missing predictions.}
\label{res_visual2}
\end{figure*}
%----------------------------------------
\begin{figure*}[h]
\centering
\resizebox{0.8\textwidth}{!}{\includegraphics{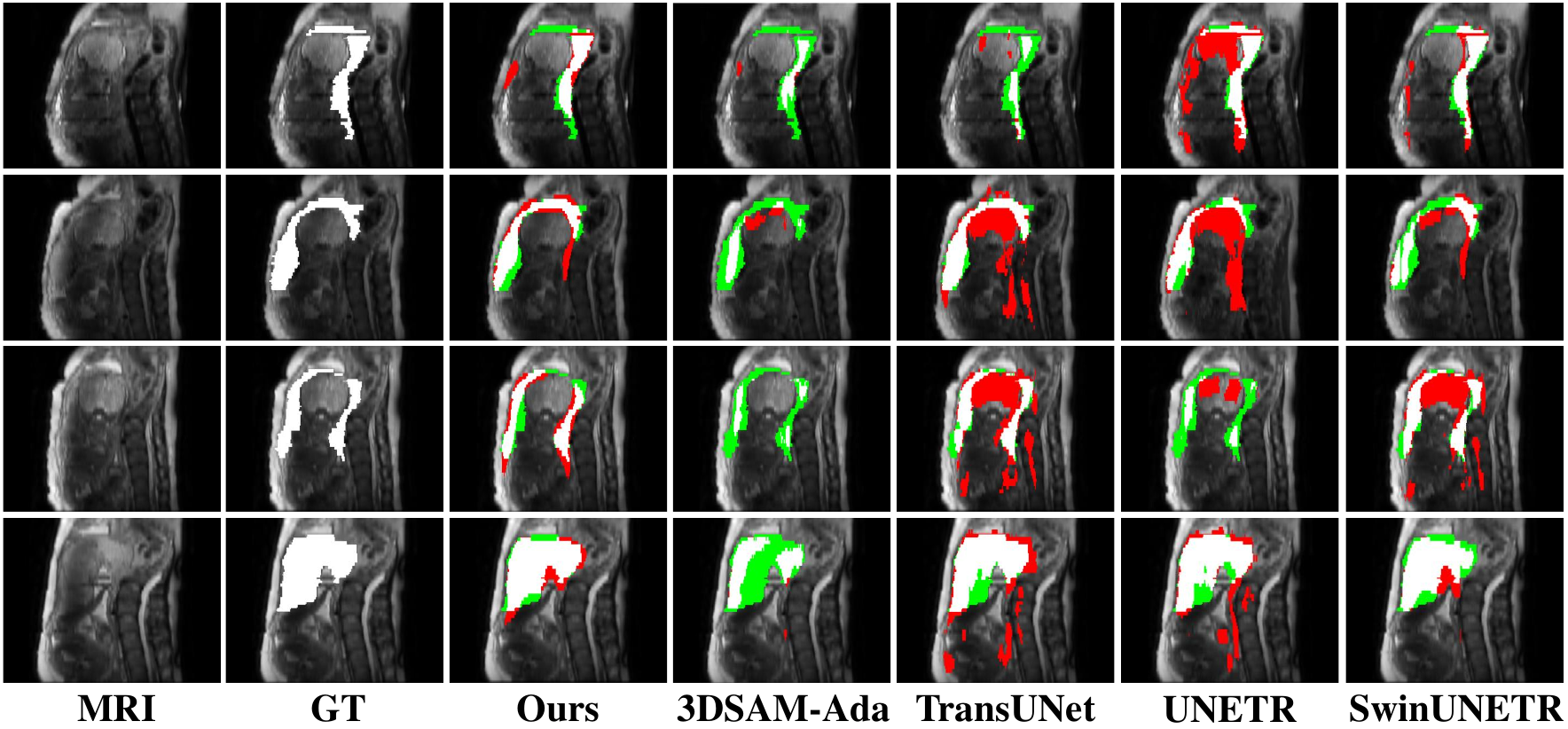}}
\centering
\caption{Visual comparison of predicted masks with different methods in the sagittal plane. The white areas indicate correctly predicted regions. The red areas represent redundant predictions, while the green areas show missing predictions.}
\label{res_visual3}
\end{figure*}
\subsubsection{Qualitative Results}
To better illustrate the superiority of our method, we present visual comparisons with different segmentation models in Fig.~\ref{res_visual1}, Fig.~\ref{res_visual2} and Fig.~\ref{res_visual3}.
From the first to the fourth row, one can observe that our method consistently demonstrates stable and robust performance across various anatomical levels.
This indicates that our method not only excels in isolated regions, but also possesses a strong overall capacity for structural perception.
Qualitative results from all three planes confirm the advantage of our method in segmenting complex anatomical structures, fusing multi-scale spatial information, and identifying lesion areas accurately.
%----------------------------------------
\begin{figure*}[h]
\centering
\resizebox{0.8\textwidth}{!}{\includegraphics{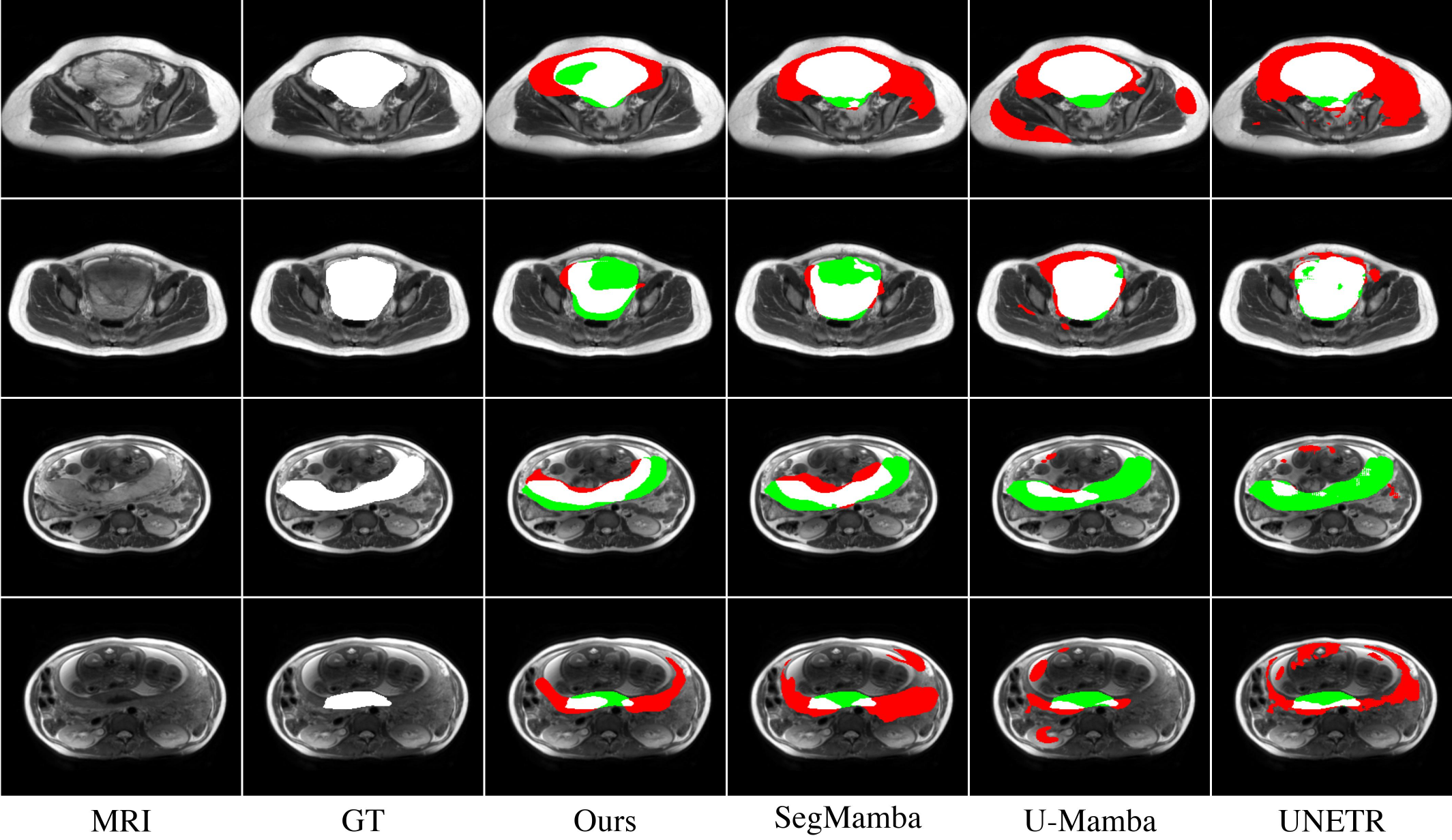}}
\centering
\caption{Representative failure cases on the PAS test set. The white areas indicate correctly predicted regions. The red areas represent redundant predictions, while the green areas show missing predictions.}
\label{res_failure}
\end{figure*}
Fig.~\ref{res_failure} presents four failure cases of 3DSAMba, including two PAS-positive cases and two PAS-negative cases.
In these samples, our method exhibits visible under-segmentation, over-segmentation, or boundary deviations, indicating that they are indeed challenging for 3DSAMba.
Nevertheless, compared with recent baselines such as SegMamba, U-Mamba, and UNETR, our method still preserves lesion completeness more effectively and produces cleaner boundaries with fewer spurious predictions.
This observation suggests that the anatomical ambiguity affecting our method also affects competing baselines, while the proposed modules enable relatively more robust lesion delineation under difficult clinical scenarios.
\subsection{Ablation Studies}
In this subsection, we explore the effectiveness of each module within our method.
We conduct experiments on the PAS dataset and observe similar results on the KiTS19 dataset.

\begin{figure}[h]
\centering
\resizebox{0.48\textwidth}{!}{\includegraphics{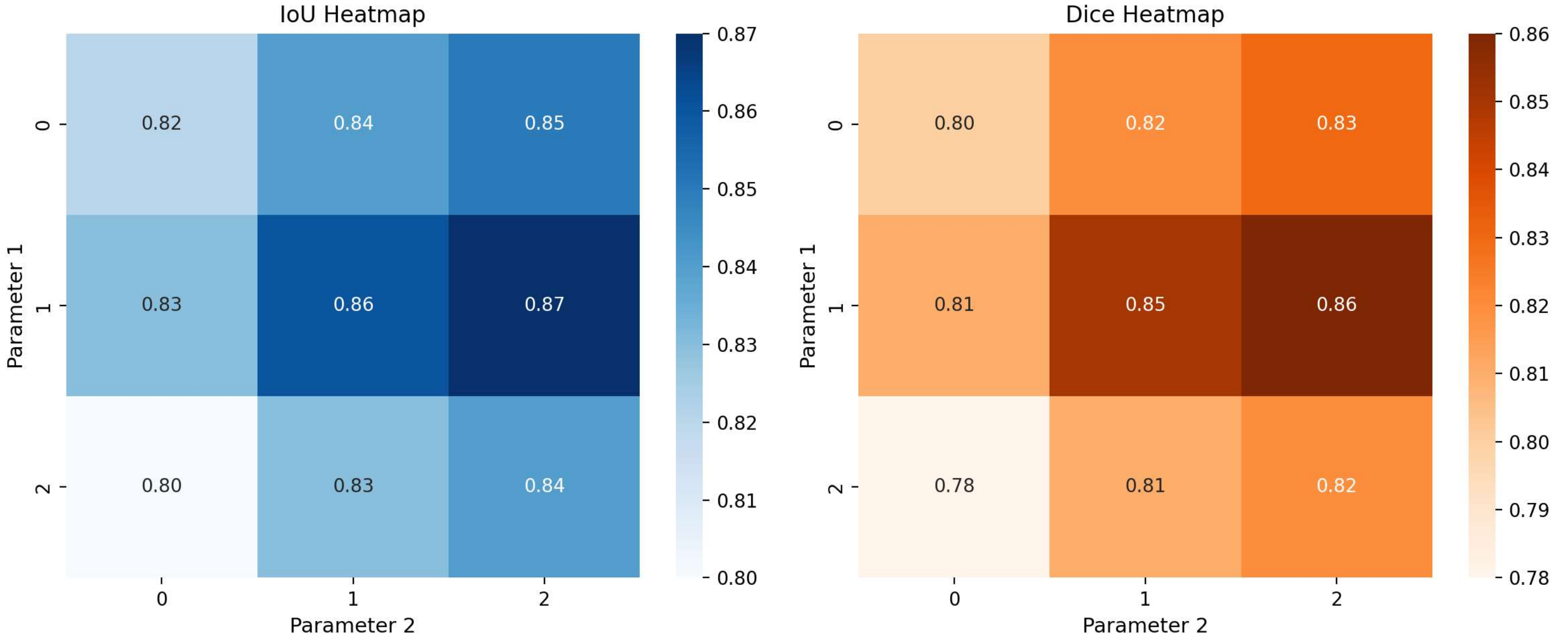}}
\centering
\caption{Performance with different hyper-parameters for lesion segmentation.}
\label{abla_hyper}
\end{figure}

\textbf{Effects of Different Hyper-parameters.}
Fig.~\ref{abla_hyper} illustrates the effects of varying the momentum and weight decay on model performance.
The left part shows IoU scores, and the right shows Dice scores.
The momentum values of 0.8, 0.9, and 0.99 are compared against the weight decay values of 0.01, 0.1, and 0.2.
The combination of momentum = 0.9 and weight decay = 0.1 yields the best performance in both metrics.
This suggests that a moderate balance of regularization and momentum is optimal.
%--------------------------------------
\begin{table}[h]
    \renewcommand\arraystretch{1.1}
    \setlength\tabcolsep{5.5pt}
    \centering
    \renewcommand\arraystretch{1.1}
    \setlength\tabcolsep{5.5pt}
    \vspace{-2mm}
    \caption{Performance with different modules on PAS dataset.}
    \resizebox{0.46\textwidth}{!}
    {
    \begin{tabular}{c|cccc|c|c}
        \hline
        &\multicolumn{4}{c|}{\textbf{Modules}} &\multicolumn{2}{c}{\textbf{PAS}}\\ \cline{2-7}
        & \textbf{Adapter}& \textbf{DSCM}& \textbf{FSSM} & \textbf{MLAM} & \textbf{Dice} & \textbf{IoU} \\
        \hline
        (A) & \ding{53}& \ding{53}& \ding{53}& \ding{53} & 0.247 & 0.153 \\
        (B) & \ding{51}& \ding{53}& \ding{53}& \ding{53} & 0.496 & 0.340 \\
        (C) & \ding{51}& \ding{51}& \ding{53}& \ding{53} & 0.677 & 0.528 \\
        (D) & \ding{51}& \ding{51}& \ding{51}& \ding{53} & 0.703 & 0.554 \\
        (E) & \ding{51}& \ding{51}& \ding{51}& \ding{51} & 0.728 & 0.580 \\
        \hline
    \end{tabular}
    }
    \label{ablation}
\end{table}
%--------------------------------------
\begin{table}[h!]
\centering
\caption{Performance comparison of fully fine-tuning and adapters.}
\small
\resizebox{0.3\textwidth}{!}
{
\begin{tabular}{l|c|c}
\hline
Model &\textbf{Dice} & \textbf{IoU}\\
\hline
Fully Fine-tuning& 0.607 & 0.445\\
Freeze+Adapters &0.677&0.528\\
\hline
\end{tabular}
}
\label{finetune}
\end{table}
%--------------------------------------
\begin{figure}[h]
\centering
\resizebox{0.46\textwidth}{!}{\includegraphics{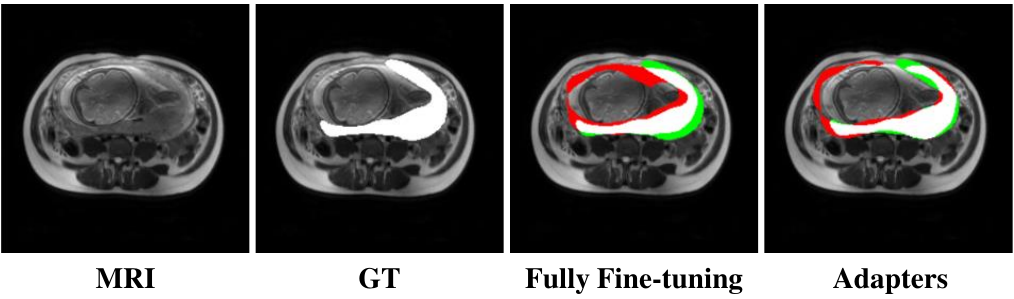}}
\centering
\caption{Visualization comparison between fully fine-tuning and adapters.}
\label{abla_fintune}
\end{figure}
%------------------------------------
\begin{figure}[h]
\centering
\resizebox{0.40\textwidth}{!}{\includegraphics{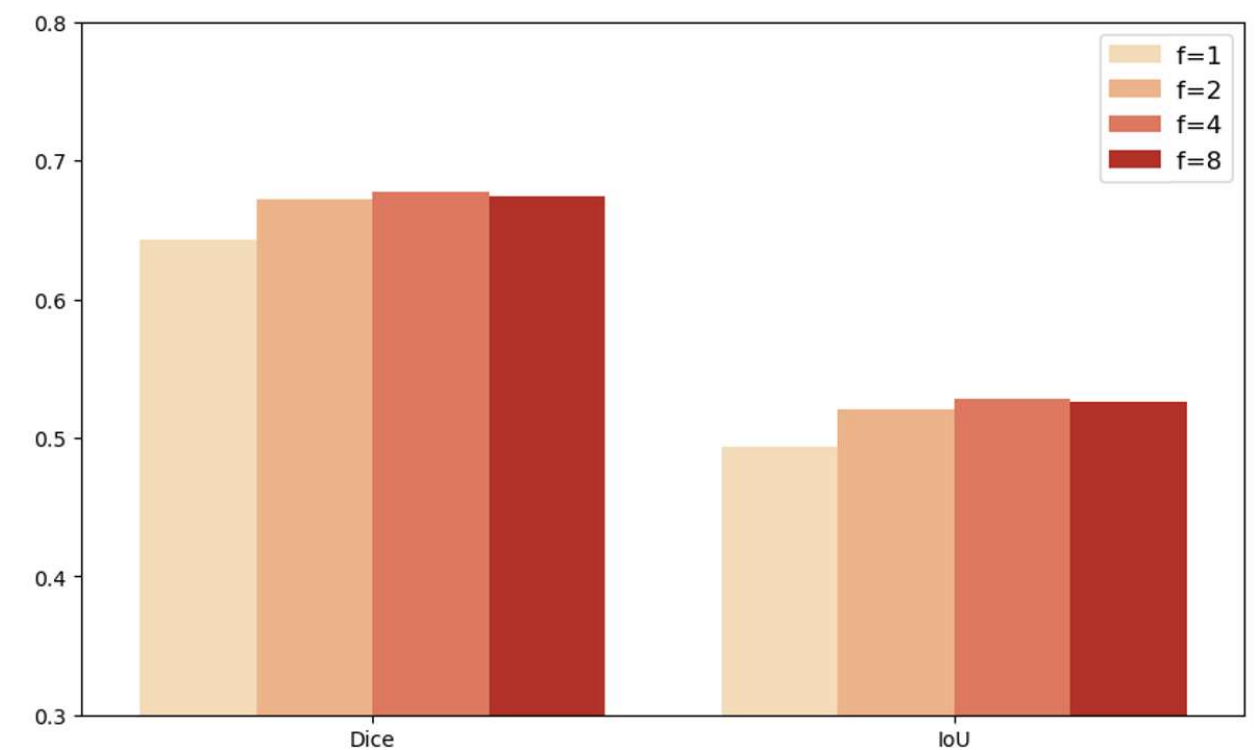}}
\centering
\caption{Performance with different projection dimensions in adapters.}
\label{abla_trainable}
\end{figure}
%---------------------------------------

\textbf{Effects of DSCM.}
In Tab.~\ref{ablation}, we compare the performance of our method with and without DSCM.
By examining rows (B) and (C), the results clearly show that the inclusion of DSCM has a significant impact on the performance.
By transforming the latent features of 3D data into a format similar to 2D features, we have significantly enhanced SAM’s ability to process 3D data.
This module allows us to maximize the retention of SAM’s original structure and parameters.

\textbf{Effects of Adapters.}
In Tab.~\ref{finetune}, we investigate the effects of fine-tuning SAM.
The first method is fine-tuning the entire SAM.
The second method freezes the SAM backbone and introduces only a small number of learnable parameters through an adapter mechanism.
The quantitative results in Tab.~\ref{finetune} and qualitative results in Fig.~\ref{abla_fintune} show that using adapters is a more effective method.
This is because the adapter mechanism allows for an efficient way to incorporate task-specific information.
In contrast, fine-tuning all parameters in a data-scarce medical scenario like PAS diagnosis has a significant instability.
In addition, freezing the SAM backbone helps better retain the model's powerful understanding abilities.

\textbf{Effects of Projection Dimensions.}
Fig.~\ref{abla_trainable} shows the effects of different projection dimensions in adapters.
First, we observe that the experimental results with varying trainable parameters are stable and robust.
Especially, the best performance is achieved when the number of projection dimensions is set to 4.
Based on these findings, we select 4 as the default number of the projection dimension.

\textbf{Effects of FSSM.}
In the rows (C) and (D) of Tab.~\ref{ablation}, we investigate the effectiveness of FSSM.
The FSSM enables an efficient fusion of features obtained from the encoder and decoder.
By scanning the sequences with SSM and calculating the channel weights, the obtained feature maps can fully integrate deep semantics and shallow details.
%------------------------------------------
\begin{figure}[h]
\centering
\includegraphics[width=0.42\textwidth,height=0.22\textwidth]{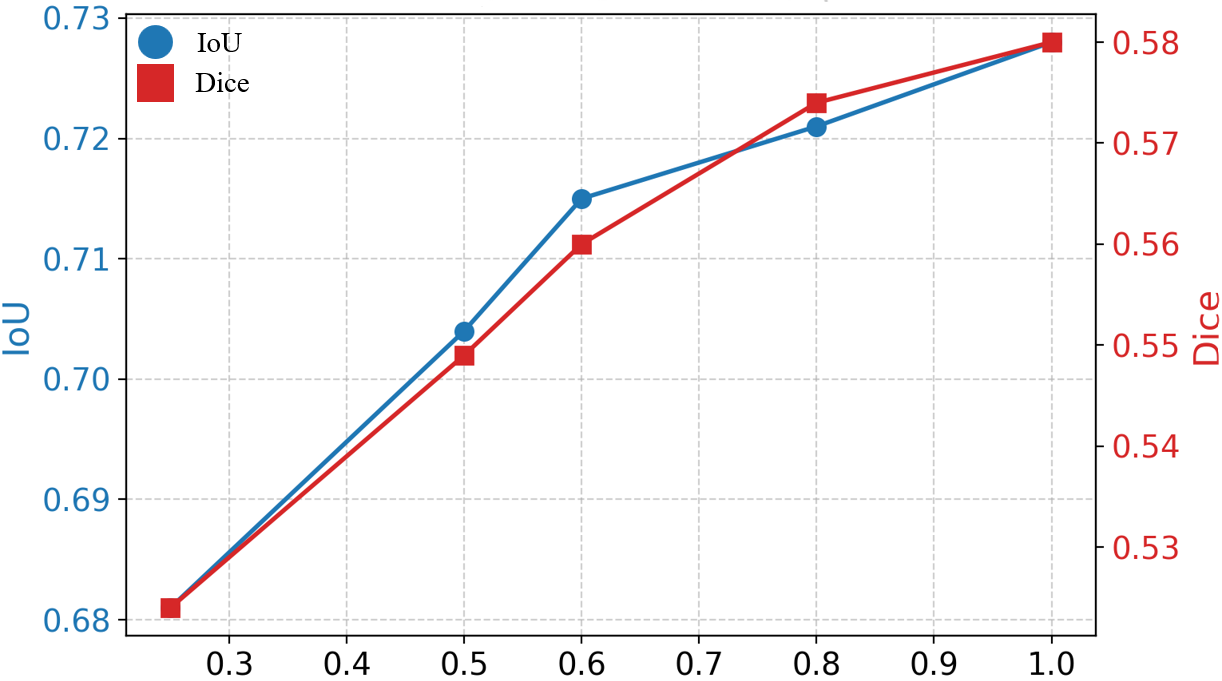}
\centering
\caption{Performance comparison with different compression rates in MLAM.}
\label{abla_squeeze_rate}
\end{figure}
%-----------------------------------------
\begin{figure}[h]
\centering
\includegraphics[width=0.42\textwidth,height=0.22\textwidth]{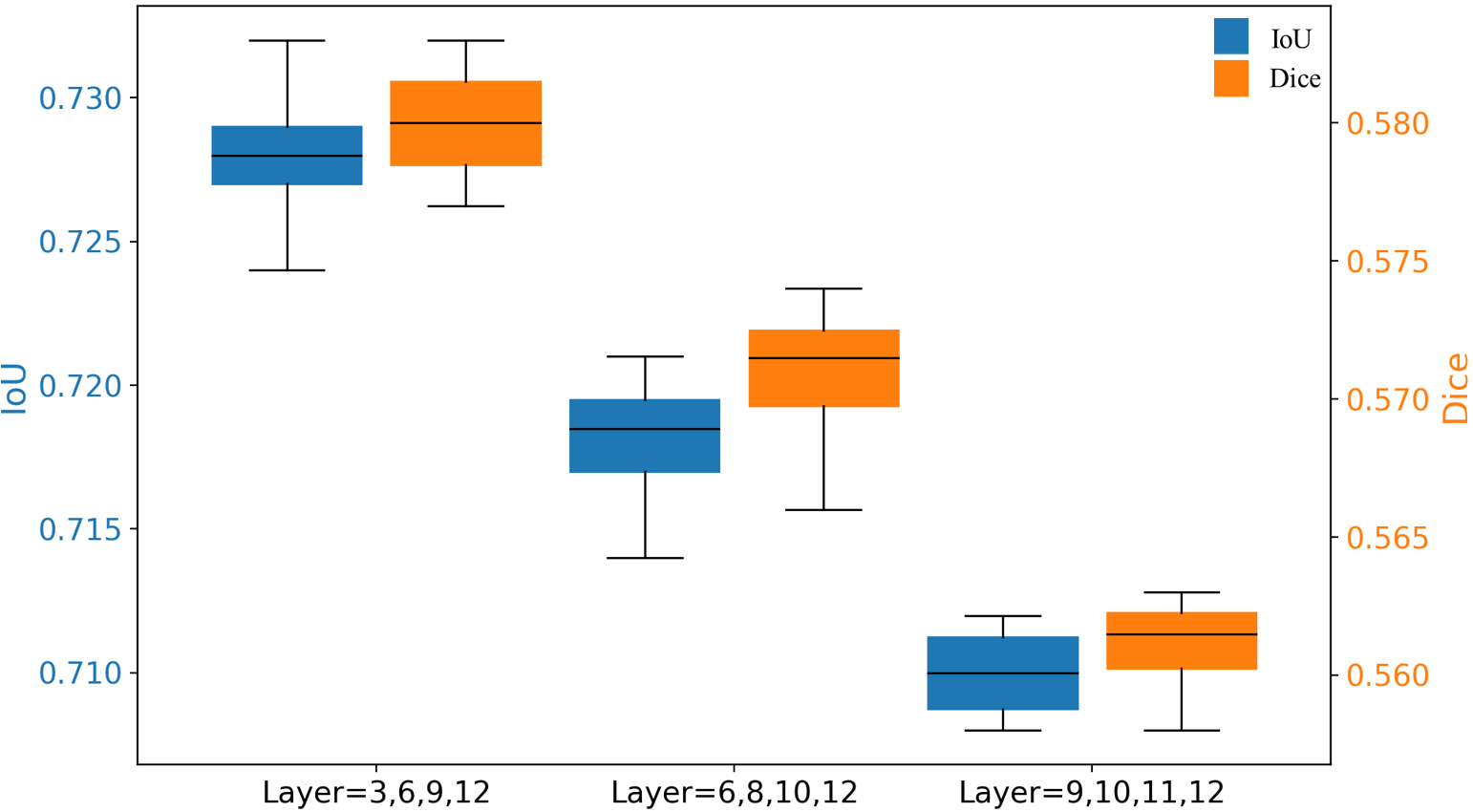}
\centering
\caption{Performance comparison with different layers in MLAM.}
\label{abla_layer}
\end{figure}
%------------------------------------------------
\begin{figure}[h]
\centering
\includegraphics[width=0.42\textwidth,height=0.18\textwidth]{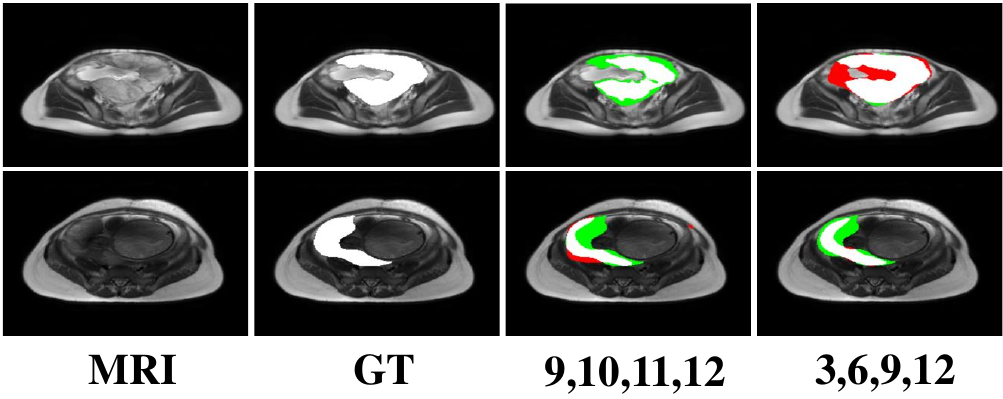}
\centering
\caption{Visual comparison with different layers used in MLAM.}
\label{abla_36912}
\end{figure}

\textbf{Effects of MLAM.}
In the rows (D) and (E) of Tab.~\ref{ablation}, we demonstrate the effectiveness of MLAM in improving the model's performance.
With MLAM, our method significantly enhances the feature representative ability of each layer.
Furthermore, Fig.~\ref{abla_squeeze_rate} shows the effect of compressing the sequence.
While the compression significantly reduces computational complexity, our experimental results show that it also substantially diminishes the model's performance.
Fig.~\ref{abla_layer} shows the effect of selected layers of SAM.
In fact, shallow layers in SAM focus more on fine-grained texture details, while deeper layers capture high-level semantic information.
When only using deep layers, the decoding process struggles to adequately restore the detailed information.
In Fig.~\ref{abla_36912}, we visualize the effects of MLAM using feature maps from different layers.
The results show that effectively integrating both shallow layers and deep layers allows for a more accurate and comprehensive segmentation.
%----------------------------------------------------
\begin{figure*}[h]
\centering
\resizebox{0.8\textwidth}{!}{\includegraphics{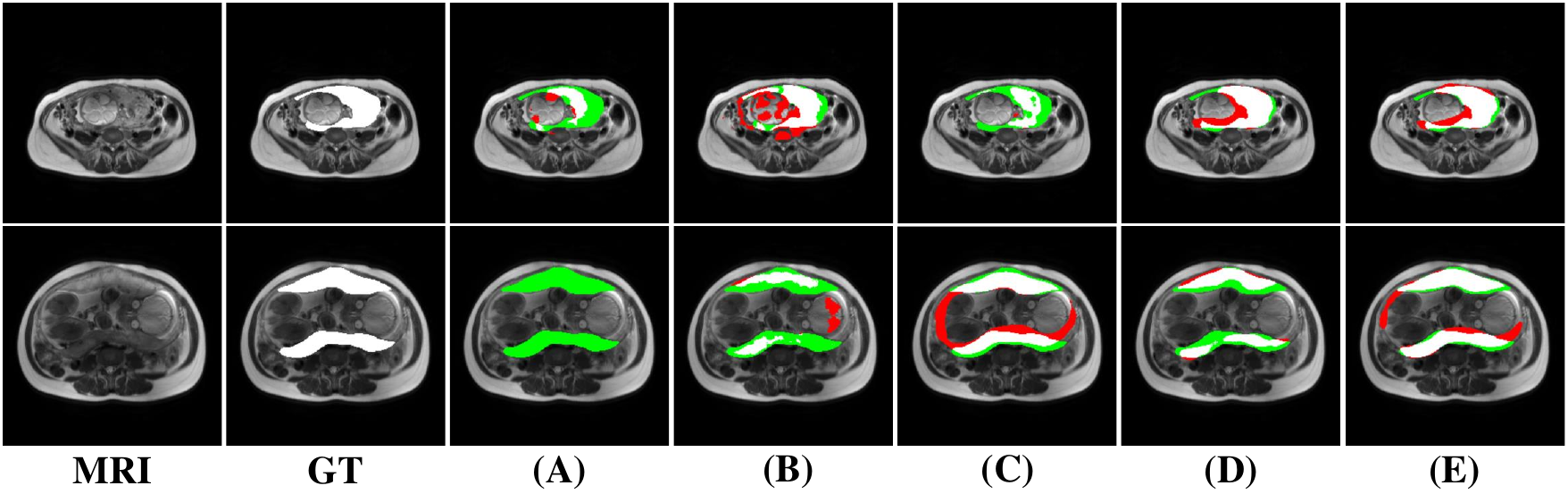}}
\centering
\caption{Visual comparison with (a) Baseline model;
(b) +Adapters;
(c) +DSCM;
(d) +FSSM;
(e) Final model.}
\label{abla_all}
\end{figure*}

\textbf{Visual Effects of Different Modules.}
Fig.~\ref{abla_all} shows the visual segmentation performance with our proposed modules.
From (A) to (E), each column represents a progressive integration of our proposed modules.
The baseline model shows evident prediction flaws, including large redundant (red) and missing (green) areas.
This indicates that the original SAM has a limited ability in identifying lesion areas and often produces imprecise boundaries and fragmented structures.
By introducing the adapter to incorporate medical domain knowledge, and utilizing DSCM to better adapt SAM to 3D data, the segmentation results become more consistent with the actual lesion shapes, with reduced redundancy and more coherent structures.
Furthermore, integrating the FSSM and MLAM significantly enhances the boundary adherence and reduces missed predictions, demonstrating a stronger structural modeling ability for accurate lesion segmentation.
Overall, the progressive improvement from (A) to (E) clearly demonstrates the crucial role of each proposed module.
%--------------------------------------------------
\begin{table}[h!]
\centering
\caption{Performance comparison of different PAS diagnosis models.}
\resizebox{0.36\textwidth}{!}{
\begin{tabular}{l|c|c}
\hline
 & Model & \textbf{OA} \\
\hline
\multirow{5}{*}{DC-based}
& NormalNet~\cite{wang2019normalnet} & 0.700 \\
& Trans3D~\cite{huang2022transformer} & 0.667 \\
& SwinTrans3D~\cite{liu2022video} & 0.667 \\
& MedMamba~\cite{yue2024medmamba} & 0.600 \\
& 3DSAMba(DC) & 0.683 \\
\hline
\multirow{6}{*}{SC-based}
& nnUNet~\cite{isensee2021nnu} & 0.750 \\
& UNETR~\cite{hatamizadeh2022unetr} & 0.767 \\
& 3DSAM-Adapter~\cite{gong20243dsam} & 0.717 \\
& U-Mamba~\cite{ma2024u} & 0.667 \\
& SegMamba~\cite{xing2024segmamba} & 0.750 \\
& 3DSAMba& \textbf{0.833} \\
\hline
\end{tabular}
}
\label{class_res}
\end{table}
%--------------------------------------------
\subsection{Diagnosis Performance of Placenta Accreta Spectrum}
To evaluate the effectiveness in diagnosing PAS, we conduct a comparative analysis of different methods.
Here, DC refers to the direct classification method, while SC denotes the method of first segmenting the lesion area and then performing classification on the segmented region.
This additional step allows the model to focus more on relevant regions.
Especially, 3DSAMba (DC) only utilizes our proposed classification network.
All SC-based methods utilize our classification network.

The results are summarized in Tab.~\ref{class_res}.
As observed, SC-based methods obviously outperform DC-based methods.
In particular, our 3DSAMba achieves the highest OA.
In general, the results confirm that the incorporation of lesion segmentation prior to classification greatly improves the accuracy of PAS diagnosis.
The superior performance of our 3DSAMba further validates the advantage of our framework in capturing subtle pathological cues for accurate lesion segmentation and reliable PAS diagnosis.
\subsection{Effect of Segmentation Mask Quality on Classification}
To further investigate how the segmentation mask quality affects the downstream classification, we analyze the relationship between the segmentation performance (Dice in Tab.~\ref{table:pas_res}) and classification accuracy (OA in Tab.~\ref{class_res}) across six SC-based methods.
A clear positive correlation can be observed: the method with the lowest Dice (U-Mamba, 0.495) also yields the lowest OA (0.667), while our 3DSAMba achieves the highest Dice (0.728) and OA (0.833).
This indicates that segmentation errors can propagate through the SC pipeline and affect the final diagnosis.
Specifically, when the predicted mask deviates from the ground truth, the classification network receives either incomplete lesion information (under-segmentation) or excessive irrelevant background (over-segmentation), both of which degrade its diagnostic capability.
Nevertheless, even the weakest SC method (U-Mamba, 0.667 OA) still outperforms most DC-based methods (e.g., MedMamba with 0.600 OA), and all other SC methods surpass the best DC method (NormalNet with 0.700 OA).
This confirms the robustness of the SC paradigm and demonstrates that even imperfect segmentation masks provide valuable guidance for PAS classification.
\subsection{Discussion on Dataset Scale}
Our PAS dataset contains 244 MRI cases (162 positive, 82 negative).
PAS is a relatively rare obstetric condition, and acquiring pelvic MRI scans with confirmed PAS diagnosis requires specialized clinical expertise and institutional resources.
Furthermore, each case in our dataset is annotated with fine-grained voxel-level segmentation masks by three experienced radiologists through a multi-round consensus process, which significantly increases the annotation cost per case compared to image-level labels.
To mitigate the limited data scale, our framework leverages the pre-trained SAM-B backbone and keeps it frozen, thereby inheriting rich visual priors from pre-training on large-scale natural images and substantially reducing the number of trainable parameters.
This design is validated by the ablation study on fully fine-tuning vs. adapters (Tab.~\ref{finetune} and Fig.~\ref{abla_fintune}), which demonstrates that freezing SAM and only training lightweight adapters yields better performance than full fine-tuning on this data-limited setting.
In addition, our method consistently outperforms ten competitive baselines across all evaluation metrics on the PAS dataset (Tab.~\ref{table:pas_res}), including both methods with fewer parameters and those with comparable or larger model sizes, suggesting that the performance gains are robust rather than artifacts of overfitting.
In the future, we plan to expand the dataset through multi-center collaboration and investigate semi-supervised or self-supervised strategies to further improve generalization.
\section{Conclusion}
In this paper, we present the first MRI-based dataset for the diagnosis of PAS.
Technically, we propose a novel feature learning framework called 3DSAMba to segment lesion areas of PAS.
Meanwhile, we adapt SAM to 3D MRI data with Deep Sequence Compression Module (DSCM) and low-rank adapters.
Furthermore, we propose the Multi-Level Aggregation Mamba (MLAM) to aggregate feature maps from different levels of the encoder.
Then, we propose the Fusion State Space Model (FSSM) to achieve an effective fusion of multi-scale features from both the encoder and decoder.
Extensive experiments demonstrate that our proposed model achieves state-of-the-art performance.
In the future, we will explore more effective methods to reduce the computation and enlarge the PAS dataset with more modalities.
% \newpage
\bibliographystyle{IEEEtran}
\bibliography{IEEEabrv,refs}

\begin{thebibliography}{10}
\providecommand{\url}[1]{#1}
\csname url@samestyle\endcsname
\providecommand{\newblock}{\relax}
\providecommand{\bibinfo}[2]{#2}
\providecommand{\BIBentrySTDinterwordspacing}{\spaceskip=0pt\relax}
\providecommand{\BIBentryALTinterwordstretchfactor}{4}
\providecommand{\BIBentryALTinterwordspacing}{\spaceskip=\fontdimen2\font plus
\BIBentryALTinterwordstretchfactor\fontdimen3\font minus
  \fontdimen4\font\relax}
\providecommand{\BIBforeignlanguage}[2]{{%
\expandafter\ifx\csname l@#1\endcsname\relax
\typeout{** WARNING: IEEEtran.bst: No hyphenation pattern has been}%
\typeout{** loaded for the language `#1'. Using the pattern for}%
\typeout{** the default language instead.}%
\else
\language=\csname l@#1\endcsname
\fi
#2}}
\providecommand{\BIBdecl}{\relax}
\BIBdecl

\bibitem{jauniaux2018placenta}
E.~Jauniaux, S.~Collins, and G.~J. Burton, ``Placenta accreta spectrum:
  pathophysiology and evidence-based anatomy for prenatal ultrasound imaging,''
  \emph{American journal of obstetrics and gynecology}, vol. 218, no.~1, pp.
  75--87, 2018.

\bibitem{poljak2023placenta}
B.~Poljak, D.~Khairudin, N.~W. Jones, and A.~K. Agten, ``Placenta accreta
  spectrum: diagnosis and management,'' \emph{Obstetrics, Gynaecology \&
  Reproductive Medicine}, vol.~33, no.~8, pp. 232--238, 2023.

\bibitem{romeo2019machine}
V.~Romeo, C.~Ricciardi, R.~Cuocolo, A.~Stanzione, F.~Verde, L.~Sarno,
  G.~Improta, P.~P. Mainenti, M.~D'Armiento, A.~Brunetti \emph{et~al.},
  ``Machine learning analysis of mri-derived texture features to predict
  placenta accreta spectrum in patients with placenta previa,'' \emph{Magnetic
  resonance imaging}, vol.~64, pp. 71--76, 2019.

\bibitem{ronneberger2015u}
O.~Ronneberger, P.~Fischer, and T.~Brox, ``U-net: Convolutional networks for
  biomedical image segmentation,'' in \emph{MICCAI}.\hskip 1em plus 0.5em minus
  0.4em\relax Springer, 2015, pp. 234--241.

\bibitem{dou20163d}
Q.~Dou, H.~Chen, Y.~Jin, L.~Yu, J.~Qin, and P.-A. Heng, ``3d deeply supervised
  network for automatic liver segmentation from ct volumes,'' in \emph{MICCAI},
  2016, pp. 149--157.

\bibitem{gibson2018automatic}
E.~Gibson, F.~Giganti, Y.~Hu, E.~Bonmati, S.~Bandula, K.~Gurusamy, B.~Davidson,
  S.~P. Pereira, M.~J. Clarkson, and D.~C. Barratt, ``Automatic multi-organ
  segmentation on abdominal ct with dense v-networks,'' \emph{TIP}, vol.~37,
  no.~8, pp. 1822--1834, 2018.

\bibitem{li2018h}
X.~Li, H.~Chen, X.~Qi, Q.~Dou, C.-W. Fu, and P.-A. Heng, ``H-denseunet: hybrid
  densely connected unet for liver and tumor segmentation from ct volumes,''
  \emph{TMI}, vol.~37, no.~12, pp. 2663--2674, 2018.

\bibitem{yu2017volumetric}
L.~Yu, X.~Yang, H.~Chen, J.~Qin, and P.~A. Heng, ``Volumetric convnets with
  mixed residual connections for automated prostate segmentation from 3d mr
  images,'' in \emph{AAAI}, vol.~31, no.~1, 2017, pp. 66--72.

\bibitem{vaswani2017attention}
A.~Vaswani, N.~Shazeer, N.~Parmar, J.~Uszkoreit, L.~Jones, A.~N. Gomez,
  {\L}.~Kaiser, and I.~Polosukhin, ``Attention is all you need,''
  \emph{NeurIPS}, vol.~30, 2017.

\bibitem{khan2022transformers}
S.~Khan, M.~Naseer, M.~Hayat, S.~W. Zamir, F.~S. Khan, and M.~Shah,
  ``Transformers in vision: A survey,'' \emph{ACM computing surveys}, vol.~54,
  no. 10s, pp. 1--41, 2022.

\bibitem{kirillov2023segment}
A.~Kirillov, E.~Mintun, N.~Ravi, H.~Mao, C.~Rolland, L.~Gustafson, T.~Xiao,
  S.~Whitehead, A.~C. Berg, W.-Y. Lo \emph{et~al.}, ``Segment anything,'' in
  \emph{ICCV}, 2023, pp. 4015--4026.

\bibitem{guefficiently}
A.~Gu, K.~Goel, and C.~Re, ``Efficiently modeling long sequences with
  structured state spaces,'' in \emph{ICLR}, 2022, pp. 1--32.

\bibitem{gu2023mamba}
A.~Gu and T.~Dao, ``Mamba: Linear-time sequence modeling with selective state
  spaces,'' \emph{arXiv}, 2023.

\bibitem{liu2024vmamba}
Y.~Liu, Y.~Tian, Y.~Zhao, H.~Yu, L.~Xie, Y.~Wang, Q.~Ye, J.~Jiao, and Y.~Liu,
  ``Vmamba: Visual state space model,'' \emph{NeurIPS}, vol.~37, pp.
  103\,031--103\,063, 2024.

\bibitem{american2018placenta}
A.~C. of~Obstetricians, Gynecologists \emph{et~al.}, ``Placenta accreta
  spectrum,'' \emph{American journal of obstetrics and gynecology}, vol. 219,
  no.~6, pp. B2--B16, 2018.

\bibitem{silver2018placenta}
R.~M. Silver and D.~W. Branch, ``Placenta accreta spectrum,'' \emph{NEJM}, vol.
  378, no.~16, pp. 1529--1536, 2018.

\bibitem{arakaza2023placenta}
A.~Arakaza, L.~Zou, and J.~Zhu, ``Placenta accreta spectrum diagnosis
  challenges and controversies in current obstetrics: a review,''
  \emph{International Journal of Women's Health}, pp. 635--654, 2023.

\bibitem{miller2021placenta}
H.~E. Miller, S.~A. Leonard, K.~A. Fox, D.~A. Carusi, and D.~J. Lyell,
  ``Placenta accreta spectrum among women with twin gestations,''
  \emph{Obstetrics \& Gynecology}, vol. 137, no.~1, pp. 132--138, 2021.

\bibitem{baughman2008placenta}
W.~C. Baughman, J.~E. Corteville, and R.~R. Shah, ``Placenta accreta: spectrum
  of us and mr imaging findings,'' \emph{Radiographics}, vol.~28, no.~7, pp.
  1905--1916, 2008.

\bibitem{happe2021predicting}
S.~K. Happe, C.~S. Yule, C.~Y. Spong, C.~E. Wells, J.~S. Dashe, E.~Moschos,
  M.~W. Rac, D.~D. McIntire, and D.~M. Twickler, ``Predicting placenta accreta
  spectrum: validation of the placenta accreta index,'' \emph{Journal of
  Ultrasound in Medicine}, vol.~40, no.~8, pp. 1523--1532, 2021.

\bibitem{srisajjakul2020magnetic}
S.~Srisajjakul, P.~Prapaisilp, and S.~Bangchokdee, ``Magnetic resonance imaging
  of placenta accreta spectrum: a step-by-step approach,'' \emph{Korean journal
  of radiology}, vol.~22, no.~2, p. 198, 2020.

\bibitem{ishibashi2020use}
H.~Ishibashi, M.~Miyamoto, H.~Shinmoto, S.~Soga, H.~Matsuura, S.~Kakimoto,
  H.~Iwahashi, T.~Sakamoto, T.~Hada, R.~Suzuki \emph{et~al.}, ``The use of
  magnetic resonance imaging to predict placenta previa with placenta accreta
  spectrum,'' \emph{Acta Obstetricia et Gynecologica Scandinavica}, vol.~99,
  no.~12, pp. 1657--1665, 2020.

\bibitem{kapoor2021review}
H.~Kapoor, M.~Hanaoka, A.~Dawkins, and A.~Khurana, ``Review of mri imaging for
  placenta accreta spectrum: pathophysiologic insights, imaging signs, and
  recent developments,'' \emph{Placenta}, vol. 104, pp. 31--39, 2021.

\bibitem{de2022diagnosis}
M.~De~Oliveira~Carniello, L.~Oliveira~Brito, L.~Sarian, and J.~Bennini,
  ``Diagnosis of placenta accreta spectrum in high-risk women using
  ultrasonography or magnetic resonance imaging: systematic review and
  meta-analysis,'' \emph{Ultrasound in Obstetrics \& Gynecology}, vol.~59,
  no.~4, pp. 428--436, 2022.

\bibitem{do2020mri}
Q.~N. Do, M.~A. Lewis, Y.~Xi, A.~J. Madhuranthakam, S.~K. Happe, J.~S. Dashe,
  R.~E. Lenkinski, A.~Khan, and D.~M. Twickler, ``Mri of the placenta accreta
  spectrum (pas) disorder: radiomics analysis correlates with surgical and
  pathological outcome,'' \emph{Journal of Magnetic Resonance Imaging},
  vol.~51, no.~3, pp. 936--946, 2020.

\bibitem{zhou2018unet++}
Z.~Zhou, M.~M. Rahman~Siddiquee, N.~Tajbakhsh, and J.~Liang, ``Unet++: A nested
  u-net architecture for medical image segmentation,'' in \emph{MICCAI}.\hskip
  1em plus 0.5em minus 0.4em\relax Springer, 2018, pp. 3--11.

\bibitem{mortazi2018automatically}
A.~Mortazi and U.~Bagci, ``Automatically designing cnn architectures for
  medical image segmentation,'' in \emph{MLMI}.\hskip 1em plus 0.5em minus
  0.4em\relax Springer, 2018, pp. 98--106.

\bibitem{chen2018drinet}
L.~Chen, P.~Bentley, K.~Mori, K.~Misawa, M.~Fujiwara, and D.~Rueckert, ``Drinet
  for medical image segmentation,'' \emph{TIP}, vol.~37, no.~11, pp.
  2453--2462, 2018.

\bibitem{chen2018encoder}
L.-C. Chen, Y.~Zhu, G.~Papandreou, F.~Schroff, and H.~Adam, ``Encoder-decoder
  with atrous separable convolution for semantic image segmentation,'' in
  \emph{ECCV}, 2018, pp. 801--818.

\bibitem{jha2019resunet++}
D.~Jha, P.~H. Smedsrud, M.~A. Riegler, D.~Johansen, T.~De~Lange, P.~Halvorsen,
  and H.~D. Johansen, ``Resunet++: An advanced architecture for medical image
  segmentation,'' in \emph{ISM}.\hskip 1em plus 0.5em minus 0.4em\relax IEEE,
  2019, pp. 225--2255.

\bibitem{gu2019net}
Z.~Gu, J.~Cheng, H.~Fu, K.~Zhou, H.~Hao, Y.~Zhao, T.~Zhang, S.~Gao, and J.~Liu,
  ``Ce-net: Context encoder network for 2d medical image segmentation,''
  \emph{TIP}, vol.~38, no.~10, pp. 2281--2292, 2019.

\bibitem{jha2020doubleu}
D.~Jha, M.~A. Riegler, D.~Johansen, P.~Halvorsen, and H.~D. Johansen,
  ``Doubleu-net: A deep convolutional neural network for medical image
  segmentation,'' in \emph{ISCMS}.\hskip 1em plus 0.5em minus 0.4em\relax IEEE,
  2020, pp. 558--564.

\bibitem{baldeon2020adaresu}
M.~Baldeon-Calisto and S.~K. Lai-Yuen, ``Adaresu-net: Multiobjective adaptive
  convolutional neural network for medical image segmentation,''
  \emph{Neurocomputing}, vol. 392, pp. 325--340, 2020.

\bibitem{adegun2021deep}
A.~A. Adegun, S.~Viriri, and R.~O. Ogundokun, ``Deep learning approach for
  medical image analysis,'' \emph{Computational Intelligence and Neuroscience},
  vol. 2021, no.~1, p. 6215281, 2021.

\bibitem{valanarasu2022unext}
J.~M.~J. Valanarasu and V.~M. Patel, ``Unext: Mlp-based rapid medical image
  segmentation network,'' in \emph{MICCAI}.\hskip 1em plus 0.5em minus
  0.4em\relax Springer, 2022, pp. 23--33.

\bibitem{alexey2020image}
D.~Alexey, ``An image is worth 16x16 words: Transformers for image recognition
  at scale,'' \emph{ICLR}, pp. 1--22, 2021.

\bibitem{liu2021swin}
Z.~Liu, Y.~Lin, Y.~Cao, H.~Hu, Y.~Wei, Z.~Zhang, S.~Lin, and B.~Guo, ``Swin
  transformer: Hierarchical vision transformer using shifted windows,'' in
  \emph{CVPR}, 2021, pp. 10\,012--10\,022.

\bibitem{xie2021cotr}
Y.~Xie, J.~Zhang, C.~Shen, and Y.~Xia, ``Cotr: Efficiently bridging cnn and
  transformer for 3d medical image segmentation,'' in \emph{MICCAI}.\hskip 1em
  plus 0.5em minus 0.4em\relax Springer, 2021, pp. 171--180.

\bibitem{chen2021transunet}
J.~Chen, Y.~Lu, Q.~Yu, X.~Luo, E.~Adeli, Y.~Wang, L.~Lu, A.~L. Yuille, and
  Y.~Zhou, ``Transunet: Transformers make strong encoders for medical image
  segmentation,'' \emph{arXiv}, 2021.

\bibitem{hatamizadeh2022unetr}
A.~Hatamizadeh, Y.~Tang, V.~Nath, D.~Yang, A.~Myronenko, B.~Landman, H.~R.
  Roth, and D.~Xu, ``Unetr: Transformers for 3d medical image segmentation,''
  in \emph{WACV}, 2022, pp. 574--584.

\bibitem{hatamizadeh2021swin}
A.~Hatamizadeh, V.~Nath, Y.~Tang, D.~Yang, H.~R. Roth, and D.~Xu, ``Swin unetr:
  Swin transformers for semantic segmentation of brain tumors in mri images,''
  in \emph{MICCAI}.\hskip 1em plus 0.5em minus 0.4em\relax Springer, 2021, pp.
  272--284.

\bibitem{chen2024ma}
C.~Chen, J.~Miao, D.~Wu, A.~Zhong, Z.~Yan, S.~Kim, J.~Hu, Z.~Liu, L.~Sun, X.~Li
  \emph{et~al.}, ``Ma-sam: Modality-agnostic sam adaptation for 3d medical
  image segmentation,'' \emph{MIA}, vol.~98, p. 103310, 2024.

\bibitem{gong20243dsam}
S.~Gong, Y.~Zhong, W.~Ma, J.~Li, Z.~Wang, J.~Zhang, P.-A. Heng, and Q.~Dou,
  ``3dsam-adapter: Holistic adaptation of sam from 2d to 3d for promptable
  tumor segmentation,'' \emph{MIA}, vol.~98, p. 103324, 2024.

\bibitem{zhang2023input}
Y.~Zhang, T.~Zhou, S.~Wang, P.~Liang, Y.~Zhang, and D.~Z. Chen, ``Input
  augmentation with sam: Boosting medical image segmentation with segmentation
  foundation model,'' in \emph{MICCAI}.\hskip 1em plus 0.5em minus 0.4em\relax
  Springer, 2023, pp. 129--139.

\bibitem{bui2024sam3d}
N.-T. Bui, D.-H. Hoang, M.-T. Tran, G.~Doretto, D.~Adjeroh, B.~Patel,
  A.~Choudhary, and N.~Le, ``Sam3d: Segment anything model in volumetric
  medical images,'' in \emph{ISBI}.\hskip 1em plus 0.5em minus 0.4em\relax
  IEEE, 2024, pp. 1--4.

\bibitem{li2023auto}
C.~Li, P.~Khanduri, Y.~Qiang, R.~I. Sultan, I.~Chetty, and D.~Zhu,
  ``Auto-prompting sam for mobile friendly 3d medical image segmentation,''
  \emph{arXiv}, 2023.

\bibitem{wang2023selective}
J.~Wang, W.~Zhu, P.~Wang, X.~Yu, L.~Liu, M.~Omar, and R.~Hamid, ``Selective
  structured state-spaces for long-form video understanding,'' in \emph{CVPR},
  2023, pp. 6387--6397.

\bibitem{ma2024u}
J.~Ma, F.~Li, and B.~Wang, ``U-mamba: Enhancing long-range dependency for
  biomedical image segmentation,'' \emph{arXiv}, 2024.

\bibitem{wang2024mamba}
Z.~Wang, J.-Q. Zheng, Y.~Zhang, G.~Cui, and L.~Li, ``Mamba-unet: Unet-like pure
  visual mamba for medical image segmentation,'' \emph{arXiv preprint
  arXiv:2402.05079}, 2024.

\bibitem{wang2024lkm}
J.~Wang, J.~Chen, D.~Chen, and J.~Wu, ``Lkm-unet: Large kernel vision mamba
  unet for medical image segmentation,'' in \emph{MICCAI}.\hskip 1em plus 0.5em
  minus 0.4em\relax Springer, 2024, pp. 360--370.

\bibitem{zhang2024vm}
M.~Zhang, Y.~Yu, S.~Jin, L.~Gu, T.~Ling, and X.~Tao, ``Vm-unet-v2: rethinking
  vision mamba unet for medical image segmentation,'' in \emph{ISBRA}.\hskip
  1em plus 0.5em minus 0.4em\relax Springer, 2024, pp. 335--346.

\bibitem{xing2024segmamba}
Z.~Xing, T.~Ye, Y.~Yang, G.~Liu, and L.~Zhu, ``Segmamba: Long-range sequential
  modeling mamba for 3d medical image segmentation,'' in \emph{MICCAI}.\hskip
  1em plus 0.5em minus 0.4em\relax Springer, 2024, pp. 578--588.

\bibitem{liu2024swin}
J.~Liu, H.~Yang, H.-Y. Zhou, L.~Yu, Y.~Liang, Y.~Yu, S.~Zhang, H.~Zheng, and
  S.~Wang, ``Swin-umamba†: Adapting mamba-based vision foundation models for
  medical image segmentation,'' \emph{TIP}, pp. 1--1, 2024.

\bibitem{wu2025h}
R.~Wu, Y.~Liu, P.~Liang, and Q.~Chang, ``H-vmunet: High-order vision mamba unet
  for medical image segmentation,'' \emph{Neurocomputing}, p. 129447, 2025.

\bibitem{shi2025frequency}
J.~Shi, T.~You, P.~Zhang, H.~Zhang, R.~Xu, and H.~Li, ``Frequency-enhanced
  multi-granularity context network for efficient vertebrae segmentation,'' in
  \emph{MICCAI}, 2025, pp. 206--216.

\bibitem{wang2026comprehensive}
C.~Wang \emph{et~al.}, ``A comprehensive analysis of mamba for 3d volumetric
  medical image segmentation,'' \emph{Pattern Recognition}, 2026.

\bibitem{ioffe2015batch}
S.~Ioffe and C.~Szegedy, ``Batch normalization: Accelerating deep network
  training by reducing internal covariate shift,'' in \emph{ICML}.\hskip 1em
  plus 0.5em minus 0.4em\relax pmlr, 2015, pp. 448--456.

\bibitem{li2017convergence}
Y.~Li and Y.~Yuan, ``Convergence analysis of two-layer neural networks with
  relu activation,'' in \emph{NeurIPS}, 2017, pp. 597--607.

\bibitem{hu2022lora}
E.~J. Hu, Y.~Shen, P.~Wallis, Z.~Allen-Zhu, Y.~Li, S.~Wang, L.~Wang, W.~Chen
  \emph{et~al.}, ``Lora: Low-rank adaptation of large language models.''
  \emph{ICLR}, vol.~1, no.~2, p.~3, 2022.

\bibitem{zhang2024fantastic}
P.~Zhang, T.~Yan, Y.~Liu, and H.~Lu, ``Fantastic animals and where to find
  them: Segment any marine animal with dual sam,'' in \emph{CVPR}, 2024, pp.
  2578--2587.

\bibitem{ba2016layer}
J.~L. Ba, J.~R. Kiros, and G.~E. Hinton, ``Layer normalization,'' \emph{STAT},
  vol. 1050, p.~21, 2016.

\bibitem{almeida2020multilayer}
L.~B. Almeida, ``Multilayer perceptrons,'' in \emph{Handbook of Neural
  Computation}.\hskip 1em plus 0.5em minus 0.4em\relax CRC Press, 2020, pp.
  C1--2.

\bibitem{nwankpa2021activation}
C.~E. Nwankpa, W.~Ijomah, A.~Gachagan, and S.~Marshall, ``Activation functions:
  comparison of trends in practice and research for deep learning,'' in
  \emph{ICCST}, 2021, pp. 124--133.

\bibitem{zhang2019depth}
R.~Zhang, F.~Zhu, J.~Liu, and G.~Liu, ``Depth-wise separable convolutions and
  multi-level pooling for an efficient spatial cnn-based steganalysis,''
  \emph{TIFS}, vol.~15, pp. 1138--1150, 2019.

\bibitem{pratiwi2020sigmoid}
H.~Pratiwi, A.~P. Windarto, S.~Susliansyah, R.~R. Aria, S.~Susilowati, L.~K.
  Rahayu, Y.~Fitriani, A.~Merdekawati, and I.~R. Rahadjeng, ``Sigmoid
  activation function in selecting the best model of artificial neural
  networks,'' in \emph{Journal of Physics: Conference Series}, vol. 1471,
  no.~1.\hskip 1em plus 0.5em minus 0.4em\relax IOP Publishing, 2020, p.
  012010.

\bibitem{heller2021state}
N.~Heller, F.~Isensee, K.~H. Maier-Hein, X.~Hou, C.~Xie, F.~Li, Y.~Nan, G.~Mu,
  Z.~Lin, M.~Han \emph{et~al.}, ``The state of the art in kidney and kidney
  tumor segmentation in contrast-enhanced ct imaging: Results of the kits19
  challenge,'' \emph{MIA}, vol.~67, p. 101821, 2021.

\bibitem{azad2024medical}
R.~Azad, E.~K. Aghdam, A.~Rauland, Y.~Jia, A.~H. Avval, A.~Bozorgpour,
  S.~Karimijafarbigloo, J.~P. Cohen, E.~Adeli, and D.~Merhof, ``Medical image
  segmentation review: The success of u-net,'' \emph{TPAMI}, pp.
  10\,076--10\,095, 2024.

\bibitem{diederik2014adam}
K.~Diederik, ``Adam: A method for stochastic optimization,'' \emph{arXiv},
  2014.

\bibitem{cciccek20163d}
{\"O}.~{\c{C}}i{\c{c}}ek, A.~Abdulkadir, S.~S. Lienkamp, T.~Brox, and
  O.~Ronneberger, ``3d u-net: learning dense volumetric segmentation from
  sparse annotation,'' in \emph{MICC}.\hskip 1em plus 0.5em minus 0.4em\relax
  Springer, 2016, pp. 424--432.

\bibitem{isensee2021nnu}
F.~Isensee, P.~F. Jaeger, S.~A. Kohl, J.~Petersen, and K.~H. Maier-Hein,
  ``nnu-net: a self-configuring method for deep learning-based biomedical image
  segmentation,'' \emph{Nature methods}, vol.~18, no.~2, pp. 203--211, 2021.

\bibitem{wang2019normalnet}
C.~Wang, M.~Cheng, F.~Sohel, M.~Bennamoun, and J.~Li, ``Normalnet: A
  voxel-based cnn for 3d object classification and retrieval,''
  \emph{Neurocomputing}, vol. 323, pp. 139--147, 2019.

\bibitem{huang2022transformer}
Y.~Huang, Y.~Si, B.~Hu, Y.~Zhang, S.~Wu, D.~Wu, and Q.~Wang,
  ``Transformer-based factorized encoder for classification of pneumoconiosis
  on 3d ct images,'' \emph{CBM}, vol. 150, p. 106137, 2022.

\bibitem{liu2022video}
Z.~Liu, J.~Ning, Y.~Cao, Y.~Wei, Z.~Zhang, S.~Lin, and H.~Hu, ``Video swin
  transformer,'' in \emph{CVPR}, 2022, pp. 3202--3211.

\bibitem{yue2024medmamba}
Y.~Yue and Z.~Li, ``Medmamba: Vision mamba for medical image classification,''
  \emph{arXiv preprint arXiv:2403.03849}, 2024.

\end{thebibliography}
\end{document}